\documentclass[runningheads]{llncs}
\usepackage{comment}

\usepackage{graphicx}
\usepackage{amsmath}
\usepackage{amssymb}
\usepackage{booktabs}

\usepackage{epsfig}
\usepackage{graphicx}
\usepackage{amsmath}
\usepackage{amssymb}
\usepackage{cite}
\usepackage{url}            % simple URL typesetting
\usepackage{booktabs}       % professional-quality tables
\usepackage{amsfonts}       % blackboard math symbols
\usepackage{amssymb}
\usepackage{nicefrac}       % compact symbols for 1/2, etc.
\usepackage{microtype}      % microtypography
\usepackage{multirow}
\usepackage{subcaption}
\usepackage{dblfloatfix}
\usepackage{textcomp}
\usepackage[ruled,lined,linesnumbered]{algorithm2e}
\usepackage{setspace}

\usepackage[square,numbers,sort&compress]{natbib}
\usepackage{enumitem}
\usepackage[font=small,labelfont=bf]{caption}
\usepackage{array}
\usepackage{multirow}
\usepackage{booktabs}
\usepackage{subcaption}
\usepackage[normalem]{ulem}
\usepackage{xparse}
\usepackage{pifont}
\usepackage{bm}
\usepackage{threeparttable}
\usepackage{etoolbox}

\usepackage{listings}
\usepackage{mwe} % for placeholder images
\usepackage{makecell}
\usepackage{color, colortbl}
\usepackage{soul}
\usepackage[scaled=0.85]{DejaVuSansMono}
\usepackage{amsmath}

\usepackage{xcolor}

\definecolor{butter1}{RGB}{252, 233,  79}
\definecolor{butter2}{RGB}{237, 212,   0}
\definecolor{butter3}{RGB}{196, 160,   0}
\colorlet{LightButter}{butter1}
\colorlet{Butter}{butter2}
\colorlet{DarkButter}{butter3}

\definecolor{orange1}{RGB}{252, 175,  62}
\definecolor{orange2}{RGB}{245, 121,   0}
\definecolor{orange3}{RGB}{206,  92,   0}
\colorlet{LightOrange}{orange1}
\colorlet{Orange}{orange2}
\colorlet{DarkOrange}{orange3}

\definecolor{chocolate1}{RGB}{233, 185, 110}
\definecolor{chocolate2}{RGB}{193, 125,  17}
\definecolor{chocolate3}{RGB}{143,  89,   2}
\colorlet{LightChocolate}{chocolate1}
\colorlet{Chocolate}{chocolate2}
\colorlet{DarkChocolate}{chocolate3}

\definecolor{chameleon1}{RGB}{138, 226,  52}
\definecolor{chameleon2}{RGB}{115, 210,  22}
\definecolor{chameleon3}{RGB}{ 78, 154,   6}
\colorlet{LightChameleon}{chameleon1}
\colorlet{Chameleon}{chameleon2}
\colorlet{DarkChameleon}{chameleon3}

\definecolor{skyblue1}{RGB}{114, 159, 207}
\definecolor{skyblue2}{RGB}{ 52, 101, 164}
\definecolor{skyblue3}{RGB}{ 32,  74, 135}
\colorlet{LightSkyBlue}{skyblue1}
\colorlet{SkyBlue}{skyblue2}
\colorlet{DarkSkyBlue}{skyblue3}

\definecolor{plum1}{RGB}{173, 127, 168}
\definecolor{plum2}{RGB}{117,  80, 123}
\definecolor{plum3}{RGB}{ 92,  53, 102}
\colorlet{LightPlum}{plum1}
\colorlet{Plum}{plum2}
\colorlet{DarkPlum}{plum3}

\definecolor{scarletred1}{RGB}{239,  41,  41}
\definecolor{scarletred2}{RGB}{204,   0,   0}
\definecolor{scarletred3}{RGB}{164,   0,   0}
\colorlet{LightScarletRed}{scarletred1}
\colorlet{ScarletRed}{scarletred2}
\colorlet{DarkScarletRed}{scarletred3}

\definecolor{aluminium1}{RGB}{238, 238, 236}
\definecolor{aluminium2}{RGB}{211, 215, 207}
\definecolor{aluminium3}{RGB}{186, 189, 182}
\definecolor{aluminium4}{RGB}{136, 138, 133}
\definecolor{aluminium5}{RGB}{ 85,  87,  83}
\definecolor{aluminium6}{RGB}{ 46,  52,  54}

\definecolor{indigo}{RGB}{114,  33, 188}
\definecolor{maroon}{RGB}{103,   7,  72}
\definecolor{turquoise}{RGB}{ 64, 224, 208}
\definecolor{green4}{RGB}{  0, 139,   0}

\newcommand{\ctext}[3][RGB]{%
  \begingroup
  \definecolor{hlcolor}{#1}{#2}\sethlcolor{hlcolor}%
  \hl{#3}%
  \endgroup
}

\newcommand{\lblsec}[1]{\label{sec:#1}}
\newcommand{\lblfig}[1]{\label{fig:#1}}
\newcommand{\lbltab}[1]{\label{tbl:#1}}
\newcommand{\lbltbl}[1]{\label{tbl:#1}}

\newcommand{\reffig}[1]{Figure~\ref{fig:#1}}
\newcommand{\reftab}[1]{Table~\ref{tbl:#1}}
\newcommand{\reftbl}[1]{Table~\ref{tbl:#1}}

\newcommand{\supplement}[1]{\cref{#1}}

\definecolor{citecolor}{HTML}{0071bc}
\usepackage[pagebackref,breaklinks=true,letterpaper=true,colorlinks,citecolor=citecolor,bookmarks=false]{hyperref}

\usepackage[capitalize]{cleveref}
\crefname{section}{\S}{\S\S}
\crefname{subsection}{\S}{\S\S}
\crefformat{table}{Table~#2#1#3}
\crefformat{figure}{Fig.~#2#1#3}
\crefformat{equation}{Eq.~#2#1#3}
\newcommand{\rowNumber}[1]{}
\definecolor{highlightRowColor}{rgb}{0.95, 0.95, 1}

\newcommand{\bI}{\mathbf{I}}
\newcommand{\bb}{\mathbf{b}}
\newcommand{\bB}{\mathbf{B}}
\newcommand{\bF}{\mathbf{F}}
\newcommand{\bof}{\mathbf{f}}
\newcommand{\bW}{\mathbf{W}}

\newcommand{\bS}{\mathbf{S}}

\newcommand{\cSup}{\mathcal{C}^{\mathrm{det}}}
\newcommand{\dSup}{\mathcal{D}^{\mathrm{det}}}
\newcommand{\cWeak}{\mathcal{C}^{\mathrm{cls}}}
\newcommand{\dWeak}{\mathcal{D}^{\mathrm{cls}}}
\newcommand{\cTest}{\mathcal{C}^{\mathrm{test}}}
\newcommand{\mAPr}{mAP$_{\text{r}}$}
\newcommand{\mAPnoval}{mAP$_{\text{novel}}$}
\newcommand{\mask}{^{\text{mask}}}
\newcommand{\bbox}{^{\text{box}}}
\newcommand{\etal}{\emph{et al.}}
\newcommand{\vs}{\emph{vs}\xspace}

\newcommand{\OURSFull}{Detector with Image Classes\xspace}
\newcommand{\OURS}{Detic\xspace}

\newcommand{\baselineDet}{Box-Supervised\xspace}

\newcommand{\vild}{ViLD\xspace}

\newcommand{\swinB}{Swin-B\xspace}

\newcommand{\imnet}{IN-21K\xspace}
\newcommand{\imnetFull}{ImageNet-21K\xspace}
\newcommand{\lvis}{LVIS\xspace}
\newcommand{\imnetLvis}{IN-L\xspace}
\newcommand{\lvisall}{LVIS-all\xspace}
\newcommand{\lvisbase}{LVIS-base\xspace}

\newcommand{\ccFull}{Conceptual Captions\xspace}
\newcommand{\cc}{CC\xspace}

\newcommand{\zeroshot}{open-vocabulary\xspace}
\newcommand{\Zeroshot}{Open-vocabulary\xspace}
\newcommand{\captionSpace}{-3mm}

\begin{document}

\pagestyle{headings}
\mainmatter
\def\ECCVSubNumber{}  % Insert your submission number here

\title{Detecting Twenty-thousand Classes \\ using Image-level Supervision}

\titlerunning{Detecting Twenty-thousand Classes}

\author{
  Xingyi Zhou$^{1,2}$
  \thanks{Work done during an internship at Meta.} 
  \quad \quad Rohit Girdhar$^1$ \quad \quad Armand Joulin$^1$ \\ Philipp Kr\"{a}henb\"{u}hl$^2$ \quad \quad Ishan Misra$^1$
}
\authorrunning{X. Zhou et al.}
\institute{$^1$Meta AI \quad \quad $^2$The University of Texas at Austin}
%******************
\maketitle

\begin{abstract}
  Current object detectors are limited in vocabulary size due to the small scale of detection datasets.
  Image classifiers, on the other hand, reason about much larger vocabularies, as their datasets are larger and easier to collect.
  We propose {\em \OURS}, which simply trains the classifiers of a detector on image classification data and thus expands the vocabulary of detectors to tens of thousands of concepts.
  Unlike prior work, \OURS does not need complex assignment schemes to assign image labels to boxes based on model predictions,
  making it much easier to implement and compatible with a range of detection architectures and backbones.
  Our results show that \OURS yields excellent detectors even for classes without box annotations.
  It outperforms prior work on both open-vocabulary and long-tail detection benchmarks.
  \OURS provides a gain of 2.4 mAP for all classes and 8.3 mAP for novel classes on the open-vocabulary LVIS benchmark.
  On the standard LVIS benchmark, \OURS obtains 41.7 mAP when evaluated on all classes, or only rare classes, hence closing the gap in performance for object categories with few samples.
  For the first time, we train a detector with all the twenty-one-thousand classes of the ImageNet dataset and show that it generalizes to new datasets without finetuning.
  Code is available at \url{https://github.com/facebookresearch/Detic}.
\end{abstract}

\section{Introduction}

Object detection consists of two sub-problems - finding the object (localization) and naming it (classification).
Traditional methods tightly couple these two sub-problems
and thus rely on box labels for all classes.
Despite many data collection efforts, detection datasets~\cite{lin2014microsoft,gupta2019lvis,shao2019objects365,kuznetsova2020open} are much smaller in overall size and vocabularies than classification datasets~\cite{deng2009imagenet}.
For example, the recent \lvis{} detection dataset~\cite{gupta2019lvis} has 1000+ classes with 120K images;
OpenImages~\cite{kuznetsova2020open} has 500 classes in 1.8M images.
Moreover, not all classes contain sufficient annotations to train a robust detector
(see \reffig{teaser} Top).
In classification, even the ten-year-old ImageNet~\cite{deng2009imagenet} has 21K classes and 14M images (\reffig{teaser} Bottom).

\begin{figure}[!t]
    \centering
    \begin{subfigure}{0.7\linewidth}
        \includegraphics[page=1, width=\linewidth]{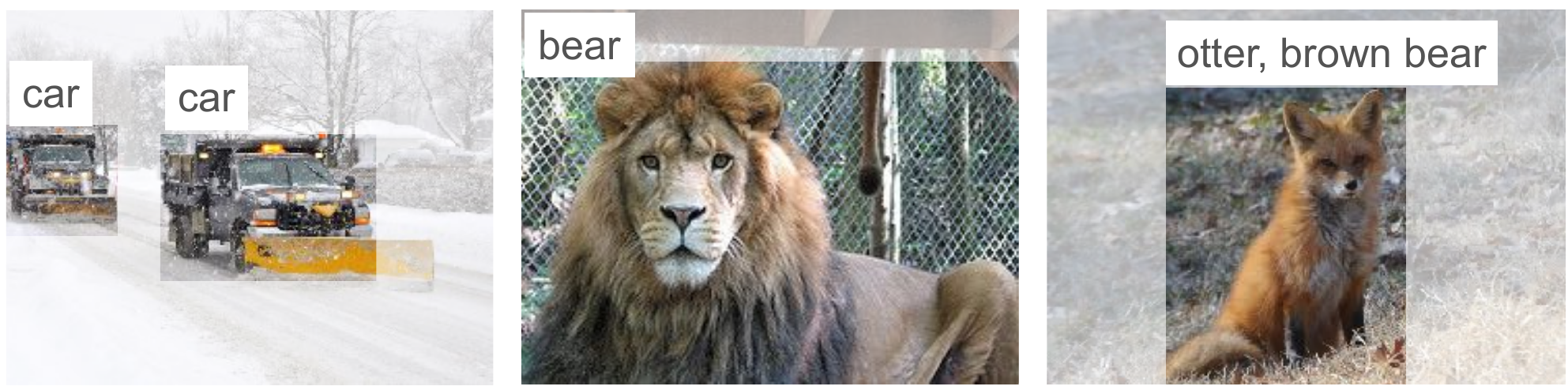}
        \vspace{-3mm}
        \end{subfigure}
    \begin{subfigure}{0.7\linewidth}
        \includegraphics[page=2, trim={0 0 2cm 0}, clip, width=\linewidth]{figs/teaser-v2-2.pdf}
        \end{subfigure}
        \vspace{\captionSpace}
       \caption{
        {\bf Top:} Typical detection results from a strong open-vocabulary LVIS detector.
        The detector misses objects of ``common'' classes.
       {\bf Bottom:} Number of images in LVIS, ImageNet, and Conceptual Captions per class (smoothed by averaging 100 neighboring classes). Classification datasets have a much larger vocabulary than detection datasets.
       }
    \lblfig{teaser}
    \vspace{-8mm}
\end{figure}

In this paper, we propose \textbf{Det}ector with \textbf{i}mage \textbf{c}lasses (\OURS) that uses
image-level supervision in addition to detection supervision.
We observe that the localization and classification sub-problems can be decoupled.
Modern region proposal networks already localize many `new' objects using existing detection supervision.
Thus, we focus on the classification sub-problem and use image-level labels to train the classifier and broaden the vocabulary of the detector.
We propose a simple classification loss that
applies the image-level supervision to the proposal with the largest size,
and do not supervise other outputs for image-labeled data.
This is easy to implement and massively expands the vocabulary.

Most existing weakly-supervised detection techniques~\cite{tang2018pcl,huang2020comprehensive,fang2021wssod,xu2021end,liu2021unbiased}
use the weakly labeled data to supervise \emph{both} the localization and classification sub-problems of detection.
Since image-classification data has no box labels, these methods develop various label-to-box assignment techniques \emph{based on model predictions} to obtain supervision.
For example, YOLO9000\cite{redmon2017yolo9000} and DLWL\cite{ramanathan2020dlwl} assign the image label to proposals that have high prediction scores on the labeled class.
Unfortunately, this prediction-based assignment requires good initial detections which
leads to a chicken-and-egg problem---we need a good detector for good label assignment, but we need many boxes to train a good detector.
Our method completely side-steps the prediction-based label assignment process by supervising the classification sub-problem alone when using classification data.
This also enables our method to learn detectors for new classes which would have been impossible to predict and assign.

Experiments on the \zeroshot{} LVIS~\cite{gupta2019lvis,gu2021zero} and
the \zeroshot{} COCO~\cite{bansal2018zero} benchmarks show
that our method can significantly improve over
a strong box-supervised baseline, on both novel and base classes.
With image-level supervision from ImageNet-21K~\cite{deng2009imagenet},
our model trained without novel class detection annotations improves the baseline
by $8.3$ point and matches the performance of using full class annotations in training.
With the standard LVIS annotations, our model reaches $41.7$ mAP and $41.7$ mAP$_{\text{rare}}$, closing the gap between rare classes and all classes.
On \zeroshot{} COCO, our method outperforms the previous state-of-the-art OVR-CNN~\cite{zareian2021open} by $5$ point with the same detector and data.
Finally, we train a detector using the full ImageNet-21K with more than twenty-thousand classes.
Our detector generalizes much better to new datasets~\cite{shao2019objects365,kuznetsova2020open} with disjoint label spaces,
reaching $21.5$ mAP on Objects365 and $55.2$ mAP50 on OpenImages,
without seeing any images from the corresponding training sets. Our contributions are summarized below:
\vspace{-2mm}
\begin{itemize}
\item We identify issues and propose a simpler alternative to existing weakly-supervised detection techniques in the open-vocabulary setting.
\item Our proposed family of losses significantly improves detection performance on novel classes, closely matching the supervised performance upper bound.
\item Our detector transfers to new datasets and vocabularies without finetuning.
\item We release our code (in supplement). It is ready-to-use for open-vocabulary detection in the real world. See examples in supplement.
\end{itemize}
\vspace{-4mm}

\begin{figure*}[!t]
    \centering
        \begin{subfigure}{0.24\linewidth}
            \includegraphics[page=1, width=0.9\linewidth]{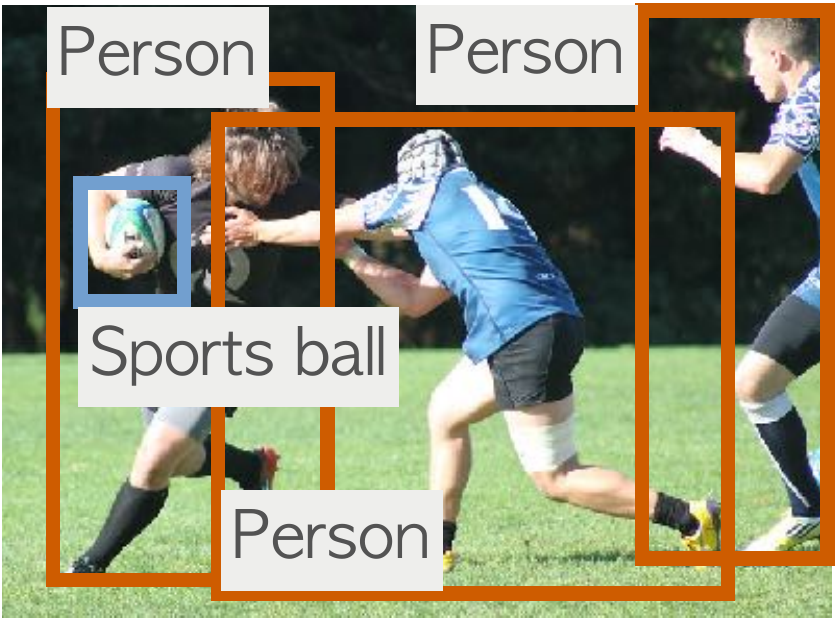}
            \caption{Standard detection}
            \end{subfigure}
            \begin{subfigure}{0.38\linewidth}
                \includegraphics[page=2, width=\linewidth]{figs/framework8_masked.pdf}
                \caption{Prediction-based label assignment}
            \end{subfigure}
            \begin{subfigure}{0.35\linewidth}
            \includegraphics[page=3, width=\linewidth]{figs/framework8_masked.pdf}
            \caption{Our non-prediction-based loss}
            \end{subfigure}
        \vspace{-3mm}
       \caption{
        \textbf{Left:} Standard detection requires ground-truth labeled boxes and cannot leverage image-level labels.
        \textbf{Center:} Existing prediction-based weakly supervised detection methods~\cite{redmon2017yolo9000,ramanathan2020dlwl,bilen2016weakly} use image-level labels by assigning them to the detector's predicted boxes (proposals).
        Unfortunately, this assignment is error-prone, especially for large vocabulary detection.
        \textbf{Right:} Detic simply assigns the image-labels to the \emph{max-size} proposal.
        We show that this loss is both simpler and performs better than prior work.
        }
    \lblfig{framework}
    \vspace{-5mm}
    \end{figure*}

\section{Related Work}

\par \noindent\textbf{Weakly-supervised object detection (WSOD)} trains object detector using image-level labels.
Many works use only image-level labels without any box supervision
~\cite{li2019weakly,shen2019cyclic,yang2019towards,wan2019c,shen2020enabling}.
WSDDN~\cite{bilen2016weakly} and OIRC~\cite{tang2017multiple} use a subnetwork to predict per-proposal weighting and sum up proposal scores into a single image scores.
PCL~\cite{tang2018pcl} first clusters proposals and then assign image labels at the cluster level.
CASD~\cite{huang2020comprehensive} further introduces feature-level attention and self-distillation.
As no bounding box supervision is used in training, these methods rely on low-level region proposal techniques~\cite{uijlings2013selective,arbelaez2014multiscale}, 
which leads to reduced localization quality.

Another line of WSOD work uses bounding box supervision together with image labels, 
known as \textbf{semi-supervised WSOD}~\cite{yan2017weakly,fang2021wssod,uijlings2018revisiting,li2018mixed,liu2021mixed,zhong2020boosting,dong2021boosting}.
YOLO9000~\cite{redmon2017yolo9000} mixes detection data and classification data in the same mini-batch, 
and assigns classification labels to anchors with the highest predicted scores.
DLWL~\cite{ramanathan2020dlwl} combines self-training and clustering-based 
WSOD~\cite{tang2018pcl}, and again assigns image labels to max-scored proposals.
MosaicOS~\cite{zhang2021mosaicos} handles domain differences between detection
and image datasets by mosaic augmentation~\cite{bochkovskiy2020yolov4}
 and proposed a three-stage self-training and finetuning framework.
In segmentation, Pinheiro \etal~\cite{pinheiro2015weakly} use a log-sum-exponential function to aggregate pixels scores into a global classification.
Our work belongs to semi-supervised WSOD.
Unlike prior work, we use a simple image-supervised loss.
Besides image labels, researchers have also studied complementary methods for
weak localization supervision like points~\cite{chen2021points} or scribles~\cite{ren2020ufo}.

\par \noindent\textbf{Open-vocabulary object detection}, or also named 
\textbf{zero-shot object detection}, aims to 
detect objects outside of the training vocabulary.
The basic solution~\cite{bansal2018zero} is to
replace the last classification layer with language embeddings 
(e.g., GloVe~\cite{pennington2014glove}) of the class names.
Rahman \etal~\cite{rahman2020improved} and Li \etal~\cite{li2019zero} improve the classifier embedding using external text information.
OVR-CNN~\cite{zareian2021open} pretrains the detector on image-text pairs.
ViLD~\cite{gu2021zero}, OpenSeg~\cite{ghiasi2021open} and langSeg~\cite{li2022language} upgrade the language embedding to CLIP~\cite{radford2021learning}.
ViLD further distills region features from CLIP image features.
We use CLIP~\cite{radford2021learning} classifier as well, but do not use distillation.
Instead, we use additional image-labeled data for co-training.

\par \noindent\textbf{Large-vocabulary object detection}~\cite{gupta2019lvis,yang2019detecting,singh2018r,redmon2017yolo9000} requires detecting 1000+ classes.
Many existing works focus on handling the long-tail problem
~\cite{pan2021model,li2020overcoming,zhang2021distribution,wu2020forest,Feng_2021_ICCV,chang2021image}.
% Repeat factor sampling (RFS)~\cite{gupta2019lvis} oversamples classes with fewer annotations.
Equalization losses~\cite{tan2020equalization,tan2021equalization} 
and SeeSaw loss~\cite{wang2021seesaw} reweights the per-class loss by balancing the gradients~\cite{tan2021equalization} or number of samples~\cite{wang2021seesaw}.
Federated Loss~\cite{zhou2021probabilistic} subsamples classes per-iteration to mimic the federated annotation~\cite{gupta2019lvis}.
Yang \etal~\cite{yang2019detecting} detects 11K classes with a label hierarchy.
Our method builds on these advances,
and we tackle the problem from a different aspect: using additional image-labeled data.

\par \noindent\textbf{Proposal Network Generalization.}
ViLD~\cite{gu2021zero} reports that region proposal networks have certain generalization abilities for new classes by default.
Dave \etal~\cite{dave2019towards} shows segmentation and localization generalizes across classes.
Kim \etal~\cite{kim2021learning} further improves proposal generalization with a localization quality estimator.
In our experiments, we found proposals to generalize well enough (see \supplement{sec:proposal}), as also observed in ViLD~\cite{gu2021zero}.
Further improvements to RPNs~\cite{gu2021zero,kim2021learning,konan2022extending,maaz2021multi} can hopefully lead to better results.

\section{Preliminaries}

We train object detectors using both object detection and image classification datasets.
We propose a simple way to leverage image supervision to learn object detectors, including for classes without box labels.
We first describe the object detection problem and then detail our approach.

\par \noindent\textbf{Problem setup.}
\lblsec{prelim}
Given an image $\bI \in \mathbb{R}^{3\times h \times w}$, object detection solves the two subproblems of (1) localization:
find all objects with their location, represented as a box $\bb_j \in \mathbb{R}^4$ and (2) classification:
assign a class label $c_j \in \cTest$ to the $j$-th object.
Here $\cTest$ is the class vocabulary provided by the user at test time.
During training, we use a detection dataset
$\dSup = \{(\bI, \{(\bb, c)_k\})_i\}_{i=1}^{|\dSup|}$ with vocabulary $\cSup$ that has both class and box labels.
We also use an image classification dataset
$\dWeak = \{(\bI, \{c_k\})_i\}_{i=1}^{|\dWeak|}$ with vocabulary $\cWeak$ that only has image-level class labels.
The vocabularies $\cTest$, $\cSup$, $\cWeak$ may or may not overlap.

\par \noindent\textbf{Traditional Object detection} considers $\cTest = \cSup$ and $\dWeak = \emptyset$.
Predominant object detectors~\cite{ren2015faster,He_2017_ICCV} follow a two-stage framework.
The first stage, called the \emph{region proposal network} (RPN), takes the image $\bI$ and produces a set of object proposals $\{(\bb, \bof, o)_j\}$, where
$\bof_j \in \mathbb{R}^D$ is a $D$-dimensional region feature and $o \in \mathbb{R}$ is the objectness score.
The second stage takes the object feature and outputs a classification score and a refined box location for each object, $s_j = \bW \bof_j$, $\hat{\bb}_j = \bB \bof_j + \bb_j$,
where $\bW \in \mathbb{R}^{|\cSup| \times D}$
and $\bB \in \mathbb{R}^{4 \times D}$
are the learned weights of the classification layer and the regression layer, respectively.\footnote{We omit the two linear layers and the bias in the second stage for notation simplicity.}
Our work focuses on improving classification in the second stage.
In our experiments,
the proposal network and the bounding box regressors are not the current performance bottleneck, as modern detectors use an over-sufficient number of proposals in testing (1K proposals for $<$ 20 objects per image. see \supplement{sec:proposal} for more details).

\par \noindent\textbf{\Zeroshot{} object detection} allows $\cTest \neq \cSup$.
Simply replacing the classification weights $\bW$ with fixed language embeddings of class
names converts a traditional detector to an \zeroshot{} detector~\cite{bansal2018zero}.
The region features are trained to match the fixed language embeddings.
We follow Gu \etal~\cite{gu2021zero} to use the CLIP embeddings~\cite{radford2021learning} as the classification weights.
In theory, this \zeroshot{} detector can detect any object class.
However, in practice, it yields unsatisfying results as shown in~\reffig{teaser}.
Our method uses image-level supervision to
improve object detection including in the \zeroshot{} setting.

\begin{figure*}[!t]
    \centering
        \begin{subfigure}{0.495\linewidth}
            \includegraphics[page=4, width=\linewidth]{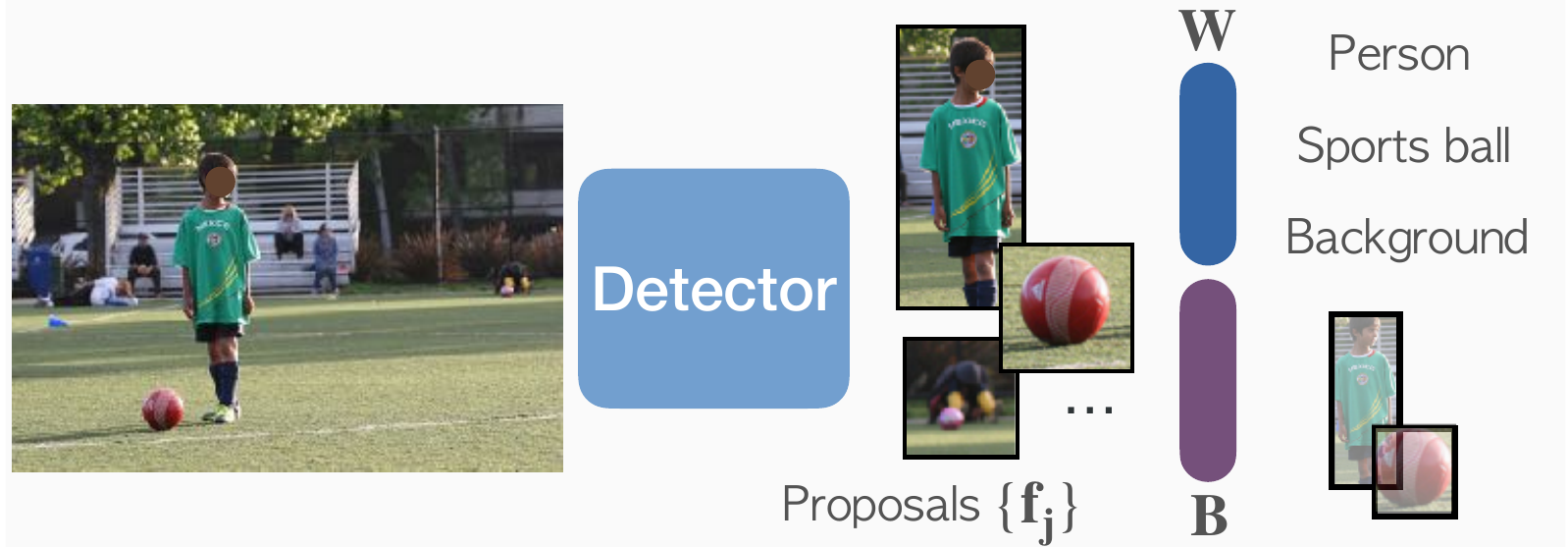}
            \caption{Detection data}
            \end{subfigure}
            \begin{subfigure}{0.47\linewidth}
                \includegraphics[page=5, width=\linewidth]{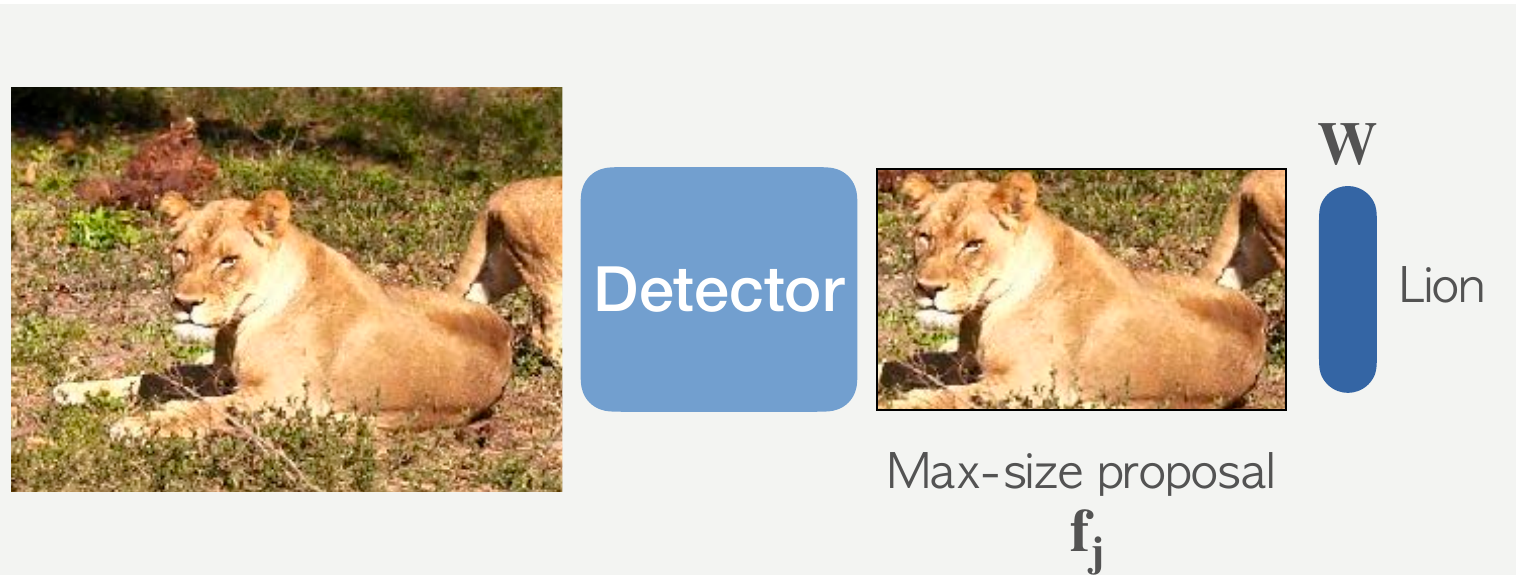}
                \caption{Image-labeled data}
            \end{subfigure}
        \vspace{\captionSpace}
       \caption{
        \textbf{Approach Overview.} We mix train on detection data and image-labeled data.
        When using detection data, our model uses the standard detection losses to train the classifier ($\mathbf{W}$) and the box prediction branch ($\mathbf{B}$) of a detector.
        When using image-labeled data, we only train the classifier using our modified classification loss.
        Our loss trains the features extracted from the largest-sized proposal.
        }
    \lblfig{framework}
    \vspace{-5mm}
\end{figure*}

\section{\OURS: \OURSFull}
\lblsec{image}
As shown in~\reffig{framework}, our method leverages the box labels from detection datasets $\dSup$ and image-level labels from classification datasets $\dWeak$.
During training, we compose a mini-batch using images from both types of datasets.
For images with box labels, we follow the standard two-stage detector training~\cite{ren2015faster}.
For image-level labeled images, we only train the features from a fixed region proposal for classification.
Thus, we only compute the localization losses (RPN loss and bounding box regression loss) on images with ground truth box labels.
Below we describe our modified classification loss for image-level labels.

A sample from the weakly labeled dataset $\dWeak$ contains an image $\bI$ and a set of $K$ labels $\{c_k\}_{k=1}^K$.
We use the region proposal network to extract $N$ object features $\{(\bb, \bof, o)_j\}_{j=1}^{N}$.
Prediction-based methods try to assign image labels to regions, and aim to train both localization and classification abilities.
Instead, we propose simple ways to use the image labels $\{c_k\}_{k=1}^K$ and only improve classification.
Our key idea is to use a fixed way to assign image labels to regions, and side-step a complex prediction-based assignment.
We allow the fixed assignment schemes miss certain objects, as long as they miss fewer objects than the prediction-based counterparts, thus leading to better performance.

\par \noindent \textbf{Non-prediction-based losses.}
We now describe a variety of simple ways to use image labels and evaluate them empirically in~\cref{tbl:imagelabel}.
Our first idea is to use the whole image as a new ``proposal'' box.
We call this loss \textbf{image-box}.
We ignore all proposals from the RPN, and instead use an injected box of the whole image $\bb' = (0, 0, w, h)$.
We then apply the classification loss to its RoI features $\bof'$ for all classes $c \in \{c_k\}_{k=1}^K$:
$$L_{\text{image-box}} = BCE(\bW \bof', c)$$
where $BCE(s, c) = -log \sigma(s_c) - \sum_{k \neq c} log (1 - \sigma(s_k))$ is the binary cross-entropy loss, and $\sigma$ is the sigmoid activation.
Thus, our loss uses the features from the same `proposal' for solving the classification problem for all the classes $\{c_k\}$.

In practice, the image-box can be replaced by smaller boxes.
We introduce two alternatives:
the proposal with the \textbf{max object score} or the proposal with the \textbf{max size}:
$$L_{\text{max-object-score}} = BCE(\bW\bof_j, c), j = \text{argmax}_j o_j$$
$$L_{\text{max-size}} = BCE(\bW\bof_j, c), j = \text{argmax}_j (\text{size}(\bb_j)) $$
We show that all these three losses can effectively leverage the image-level supervision, while the max-size loss performs the best.
We thus use the max-size loss by default for image-supervised data.
We also note that the classification parameters $\bW$ are shared across both detection and classification data, which greatly improves detection performance. The overall training objective is
\begin{equation*}
    L(\bI)=\begin{cases}
        L_{\text{rpn}} + L_{\text{reg}} + L_{\text{cls}},  & \text{if} \ \bI \in \dSup \\
        \lambda L_{\text{max-size}}, & \text{if} \ \bI \in \dWeak
    \end{cases}
\end{equation*}
where $L_{\text{rpn}}$, $L_{\text{reg}}$, $L_{\text{cls}}$ are standard losses in a two-stage detector, and $\lambda=0.1$ is the weight of our loss.

\par \noindent \textbf{Relation to prediction-based assignments.}
In traditional weakly-supervised detection
~\cite{bilen2016weakly,redmon2017yolo9000,ramanathan2020dlwl},
a popular idea is to assign the image to the proposals based on model prediction.
Let $\bF = (\bof_1, \dots, \bof_N)$ be the stacked feature of all object proposals
and $\bS = \bW\bF$ be their classification scores.
For each $c \in \{c_k\}_{k=1}^K$,
$L = BCE(\bS_j, c), j = \mathcal{F}(\bS, c)$,
where $\mathcal{F}$ is the label-to-box assignment process.
In most methods, $\mathcal{F}$ is a function of the prediction $\bS$.
For example, $\mathcal{F}$ selects the proposal with max score on $c$.
Our key insight is that $\mathcal{F}$ should \emph{not} depend on the prediction $\bS$.
In large-vocabulary detection,
the initial recognition ability of rare or novel classes is low,
making the label assignment process inaccurate.
Our method side-steps this prediction-and-assignment process entirely and relies on a fixed supervision criteria.
\section{Experiments}
\lblsec{experiment}
We evaluate \OURS{} on the large-vocabulary object detection dataset \lvis{}~\cite{gupta2019lvis}.
We mainly use the \zeroshot{} setting proposed by Gu \etal~\cite{gu2021zero}, and also report results on the standard \lvis{} setting.
We describe our experiment setup below.

\par \noindent \textbf{\lvis{}.}
The \lvis{}~\cite{gupta2019lvis} dataset has object detection and instance segmentation labels for $1203$ classes with 100K images.
The classes are divided into three groups - frequent, common, rare based on the number of training images.
We refer to this standard \lvis{} training set as \emph{\lvisall}.
Following \vild{}~\cite{gu2021zero},
we remove the labels of $337$ rare-class from training and consider them as novel classes in testing.
We refer to this partial training set with only frequent and common classes as \emph{\lvisbase}.
We report mask mAP which is the official metric for LVIS.
While our model is developed for box detection,
we use a standard class-agnostic mask head~\cite{He_2017_ICCV} to produce segmentation masks for boxes.
We train the mask head only on detection data.

\par \noindent \textbf{Image-supervised data.}
We use two sources of image-supervised data: \imnetFull~\cite{deng2009imagenet} and \ccFull~\cite{sharma2018conceptual}.
\imnetFull{} (IN-21K) contains 14M images for 21K classes.
For ease of training and evaluation, most of our experiments use the 997 classes that overlap with the \lvis{} vocabulary and denote this subset as \imnetLvis.
\ccFull~\cite{sharma2018conceptual} (CC) is
an image captioning dataset containing 3M images.
We extract image labels from the captions using exact text-matching and keep images whose captions mention at least one \lvis{} class.
See \supplement{sec:caption} for results of directly using captions.
The resulting dataset contains 1.5M images with 992 \lvis{} classes.
We summarize the datasets used below.
\begin{table}[!h]
% \small
\vspace{-8mm}
\begin{center}
\begin{tabular}{@{}l@{\ \ \ \ }c@{\ \ \ \ }c@{\ \ \ \ }c@{}}
\toprule
Notation & Definition & \#Images & \#Classes\\
\midrule
\lvisall & The original \lvis{} dataset~\cite{gupta2019lvis} & 100K & 1203 \\
\lvisbase & \lvis{} without rare-class annotations & 100K & 866 \\
\imnet & \!\!\!The original \imnetFull{} dataset~\cite{deng2009imagenet} & 14M & 21k \\
\imnetLvis &  \!\!\!\!\!\!\!997 overlapping \imnet{} classes with \lvis{} & 1.2M & 997 \\
\cc & \!\!\!\!\!\!\!\ccFull~\cite{sharma2018conceptual} with \lvis{} classes & 1.5M & 992 \\
\bottomrule
 \end{tabular}
\end{center}
\lbltab{datasets}
\vspace{-14mm}
\end{table}

\subsection{Implementation details}
\label{sec:implementation_details}

\noindent \textbf{\baselineDet: a strong \lvis{} baseline.}
We first establish a strong baseline on \lvis{} to demonstrate that our improvements are orthogonal to recent advances in object detection.
The baseline only uses the supervised bounding box labels.
We use the CenterNet2~\cite{zhou2021probabilistic} detector with ResNet50~\cite{he2016deep} backbone.
We use Federated Loss~\cite{zhou2021probabilistic} and repeat factor sampling~\cite{gupta2019lvis}.
We use large scale jittering~\cite{ghiasi2021simple} with input resolution $640\!\times\!640$ and train for a $4\times$ ($\sim\!\!48$ LVIS epochs) schedule.
To show our method is compatible with better pretraining,
we use ImageNet-21k pretrained backbone weights~\cite{ridnik2021imagenet21k}.
As described in \cref{sec:prelim}, we use the CLIP~\cite{radford2021learning} embedding as the classifier.
Our baseline is $9.1$ mAP higher than the detectron2
baseline~\cite{wu2019detectron2} ($31.5$ vs. $22.4$ mAP$\mask$) and trains in a similar time (17 vs. 12 hours on 8 V100 GPUs).
See \supplement{sec:lvis-baseline} for more details.

\noindent\textbf{Resolution change for image-labeled images.}
ImageNet images are inherently smaller and more object-focused than LVIS images~\cite{zhang2021mosaicos}.
In practice, we observe it is important to use smaller image resolution for ImageNet images.
Using smaller resolution in addition allows us to increase the batch-size with the same computation.
In our implementation, we use $320\!\!\times\!\!320$ for ImageNet and CC and ablate this in~\supplement{sec:ratio-and-size}.

\noindent\textbf{Multi-dataset training.}
We sample detection and classification mini-batches in a $1:1$ ratio,
regardless of the original dataset size.
We group images from the same dataset on the same GPU to improve training efficiency~\cite{zhou2021simple}.

\par \noindent\textbf{Training schedules.}
To shorten the experimental cycle and have a good initialization
for prediction-based WSOD losses~\cite{ramanathan2020dlwl,redmon2017yolo9000},
we always first train a converged base-class-only model ($4\times$ schedule)
and finetune on it with additional image-labeled data for another $4\times$ schedule.
We confirm finetuning the model using only box supervision does not improve the performance.
The $4\times$ schedule for our joint training consists of $\sim\!\!24$ LVIS epochs plus $\sim\!\!4.8$ ImageNet epochs or $\sim\!\!3.8$ CC epochs.
Training our ResNet50 model takes $\sim\!22$ hours on 8 V100 GPUs.
The large 21K \swinB model trains in $\sim\!24$ hours on 32 GPUs.

\subsection{Prediction-based \vs non-prediction-based methods}

\reftab{imagelabel} shows the results of the box-supervised baseline, existing prediction-based methods, and our proposed non-prediction-based methods.
The baseline (Box-Supervised) is trained without access to novel class bounding box labels.
It uses the CLIP classifier~\cite{gu2021zero} and has \zeroshot{} capabilities with 16.3 mAP$_{\text{novel}}$.
In order to leverage additional image-labeled data like ImageNet or \cc, we use prior prediction-based methods or our non-prediction-based method.

We compare a few prediction-based methods that assign image labels to proposals based on predictions.
Self-training assigns predictions of \baselineDet{} as pseudo-labels \emph{offline} with a fixed score threshold ($0.5$).
The other prediction-based methods use different losses to assign predictions to image labels online.
See~\supplement{sec:predictionbaseddetails} for implementation details.
For DLWL~\cite{ramanathan2020dlwl}, we implement a simplified version that does not include bootstrapping and refer to it as DLWL*.

\begin{table*}[!t]
    \begin{center}
    \begin{tabular}{@{}ll@{}c@{\ \ }c@{\ \ \ \ \ \ \ \ }c@{\ \ }c@{}}
    \toprule
    \rowNumber{\#} & & \multicolumn{2}{c}{\!\!\!\!IN-L (object-centric)} & \multicolumn{2}{c}{\!\!\!\!CC (non object-centric)}\\
    & & mAP$\mask$ & mAP$\mask_{\text{novel}}$ & mAP$\mask$ & mAP$\mask_{\text{novel}}$\\
    \cmidrule(r){1-2}
    \cmidrule(r){3-4}
    \cmidrule(r){5-6}
    \rowNumber{1} & Box-Supervised (baseline)
    & 30.0\tiny$\pm 0.4$ & 16.3\tiny$\pm 0.7$
                            & 30.0\tiny$\pm 0.4$ & 16.3\tiny$\pm 0.7$ \\
    \cmidrule(r){1-2}
    \cmidrule(r){3-4}
    \cmidrule(r){5-6}
    \multicolumn{5}{l}{\emph{Prediction-based methods}} \\
    \rowNumber{2} & Self-training~\cite{sohn2020simple} & 30.3\tiny$\pm 0.0$ & 15.6\tiny$\pm 0.1$
                                        & 30.1\tiny$\pm 0.2$ & 15.9\tiny$\pm 0.8$\\
    \rowNumber{3} & WSDDN~\cite{bilen2016weakly} & 29.8\tiny$\pm 0.2$ & 15.6\tiny$\pm 0.3$
                                 & 30.0\tiny$\pm 0.1$ & 16.5\tiny$\pm 0.8$\\
    \rowNumber{4} & DLWL*~\cite{ramanathan2020dlwl} & 30.6\tiny$\pm 0.1$ & 18.2\tiny$\pm 0.2$
                                    & 29.7\tiny$\pm 0.3$ & 16.9\tiny$\pm 0.6$\\
    \rowNumber{5} &
    YOLO9000~\cite{redmon2017yolo9000}
        & 31.2\tiny$\pm 0.3$ & 20.4\tiny$\pm 0.9$
        & 29.4\tiny$\pm 0.1$ & 15.9\tiny$\pm 0.6$ \\
    \cmidrule(r){1-2}
    \cmidrule(r){3-4}
    \cmidrule(r){5-6}
    \multicolumn{5}{l}{\emph{Non-prediction-based methods}} \\
    \rowNumber{6} & \OURS{} (Max-object-score) & 32.2\tiny$\pm 0.1$ & 24.4\tiny$\pm 0.3$ & 29.8\tiny$\pm 0.1$ & 18.2\tiny$\pm 0.6$ \\
    \rowNumber{7} & \OURS{} (Image-box) & \bf 32.4\tiny$\pm 0.1$& 23.8\tiny$\pm 0.5$& \bf 30.9\tiny$\pm 0.1$ & \bf 19.5\tiny$\pm 0.5$ \\
    \rowcolor{highlightRowColor} \rowNumber{8} & \OURS{} (Max-size) & \bf 32.4\tiny$\pm 0.1$ & \bf 24.6\tiny$\pm 0.3$
                   & \bf 30.9\tiny$\pm 0.2$ & \bf 19.5\tiny$\pm 0.3$ \\

    \cmidrule(r){1-2}
    \cmidrule(r){3-4}
    \cmidrule(r){5-6}
    \rowNumber{9} & \color{gray} Fully-supervised (all classes) %\color{gray} Box-Supervised (all classes)
      & \color{gray} 31.1\tiny$\pm 0.4$ & \color{gray} 25.5\tiny$\pm 0.7$
      & \color{gray} 31.1\tiny$\pm 0.4$ & \color{gray} 25.5\tiny$\pm 0.7$ \\
    \bottomrule
     \end{tabular}
    \end{center}
    \vspace{\captionSpace}
    \caption{
    \textbf{Prediction-based \vs non-prediction-based methods.}
    We show overall and novel-class mAP on \zeroshot{} LVIS~\cite{gu2021zero} (with 866 base classes and 337 novel classes) with different image-labeled datasets (IN-L or CC).
    The models are trained using our strong baseline~\cref{sec:implementation_details} (top row).
    This baseline is trained on boxes from the base classes and has non-zero novel-class mAP as it uses the CLIP classifier.
    All models in the following rows are finetuned from the baseline model and leverage image-labeled data.
    We repeat experiments for 3 runs and report mean/ std.
    All variants of our proposed non-prediction-based losses outperform existing prediction-based counterparts.
    }
    \lbltab{imagelabel}
    \vspace{-8mm}
\end{table*}

~\reftab{imagelabel} (third block) shows the results of our non-prediction-based methods in \cref{sec:image}.
All variants of our proposed simpler method outperform the complex prediction-based counterparts, with both image-supervised datasets.
On the novel classes, \OURS{} provides a significant gain of $\sim4.2$ points with ImageNet over the best prediction-based methods.

\par \noindent \textbf{Using non-object centric images from \ccFull.}
ImageNet images typically have a single large object~\cite{gupta2019lvis}.
Thus, our non-prediction-based methods, for example image-box which considers the entire image as a bounding box, are well suited for ImageNet.
To test whether our losses work with different image distributions with multiple objects, we test it with the \ccFull (\cc) dataset.
Even on this challenging dataset with multiple objects/labels per image, \OURS{} provides a gain of $\sim2.6$ points on novel class detection over the best prediction-based methods.
This suggests that our simpler \OURS{} method can generalize to different types of image-labeled data.
Overall, the results from~\cref{tbl:imagelabel} suggest that complex prediction-based methods that overly rely on model prediction scores do not perform well for open-vocabulary detection.
Amongst our non-prediction-based variants, the max-size loss consistently performs the best, and 
is the default for \OURS{} in our following experiments.

\par \noindent\textbf{Why does max-size work?}
Intuitively, our simpler non-prediction methods outperform the complex prediction-based method by side-stepping a hard assignment problem.
Prediction-based methods rely on strong initial detections to assign image-level labels to predicted boxes.
When the initial predictions are reliable, prediction-based methods are ideal.
However, in open-vocabulary scenarios, such strong initial predictions are absent, which explains the limited performance of prediction-based methods.
\OURS{}'s simpler assignment does not rely on strong predictions 
and is more robust under the challenges of open-vocabulary setting.

We now study two additional advantages of the \OURS{} max-size variant over prediction-based methods that may contribute to improved performance:
1) the selected max-size proposal can safely \emph{cover} the target object;
2) the selected max-size proposal is consistent during different training iterations.

\reffig{qualitative-proposal} provides typical qualitative examples of the assigned region for the prediction-based method and our max-size variant.
On an annotated subset of IN-L, \OURS{} max-size covers $92.8\%$ target objects, vs. $69.0\%$ for the prediction-based method.
Overall, unlike prediction-based methods, \OURS{}'s simpler assignment yields boxes that are more likely to contain the object.
Indeed, \OURS{} may miss certain objects (especially small objects) or supervise to a loose region.
However, in order for \OURS{} to yield a good detector, the selected box need not be perfect, it just needs to 1) provide meaningful training signal (cover the objects and be consistent during training); 2) be `more correct' than the box selected by the prediction-based method.
We provide details about our metrics, more quantitative evaluation, and more discussions in \supplement{sec:comparison-prediction-based}.

\begin{figure}[!t]
    \centering
    \scriptsize
          \begin{tabular}{@{}c@{\ }c@{}c@{}c@{}c@{}c@{}c@{}}
         & $1/3$ iters & $2/3$ iters & Convergence & $1/3$ iters & $2/3$ iters & Convergence \\
          \rotatebox[origin=c]{90}{\scriptsize Pred.-based\!\!\!\!\!\!\!\!\!\!\!\!\!\!\!\!\!\!\!\!\!\!\!\!\!\!\!\!\!\!} & \includegraphics[width=0.17\linewidth]{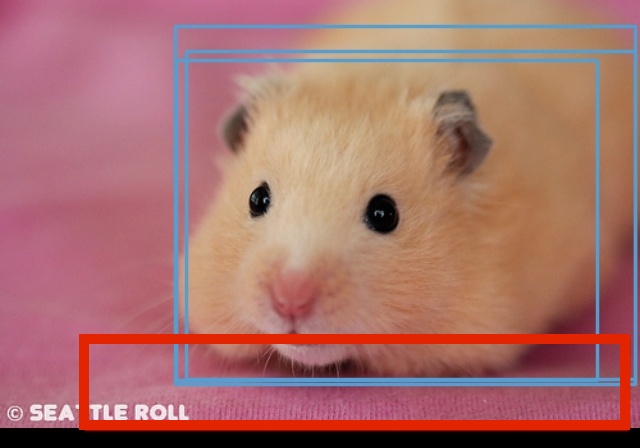}
          & \includegraphics[width=0.17\linewidth]{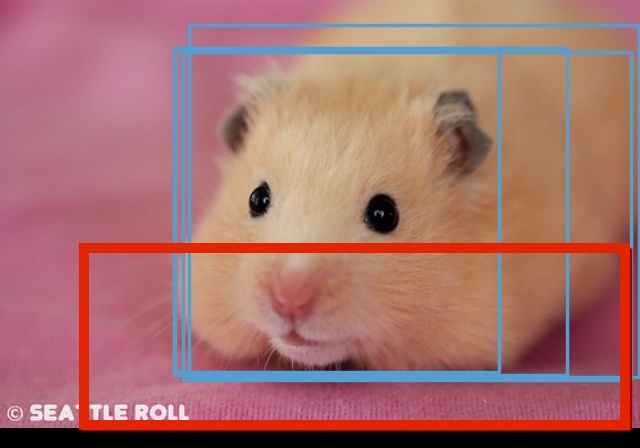}
          & \includegraphics[width=0.17\linewidth]{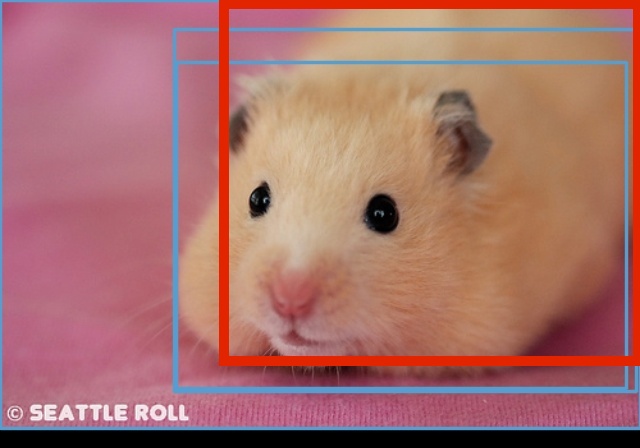}
          & \includegraphics[trim={0 1cm 0 4.0cm}, clip, width=0.15\linewidth]{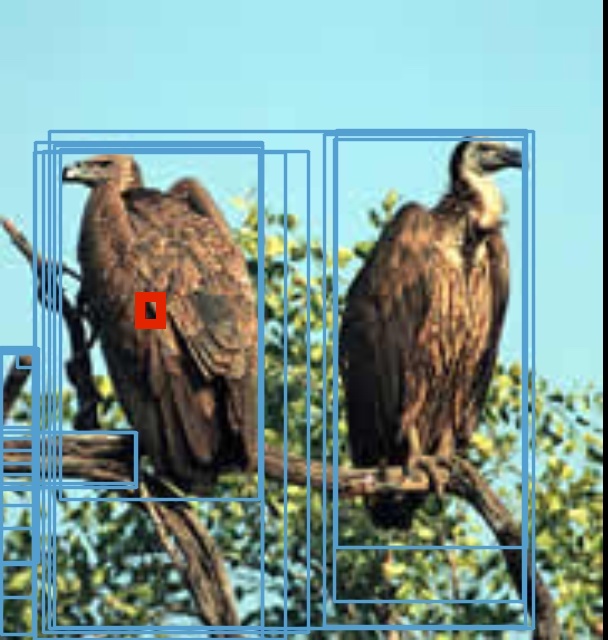}
          & \includegraphics[trim={0 1cm 0 4.0cm}, clip, width=0.15\linewidth]{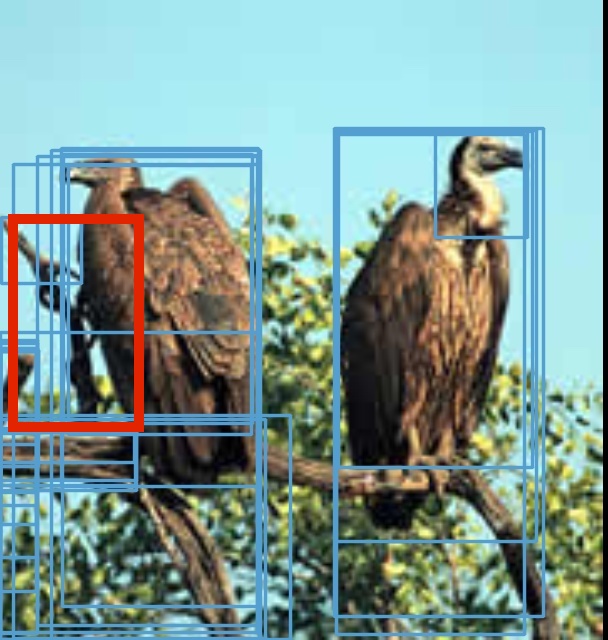}
          & \includegraphics[trim={0 1cm 0 4.0cm}, clip, width=0.15\linewidth]{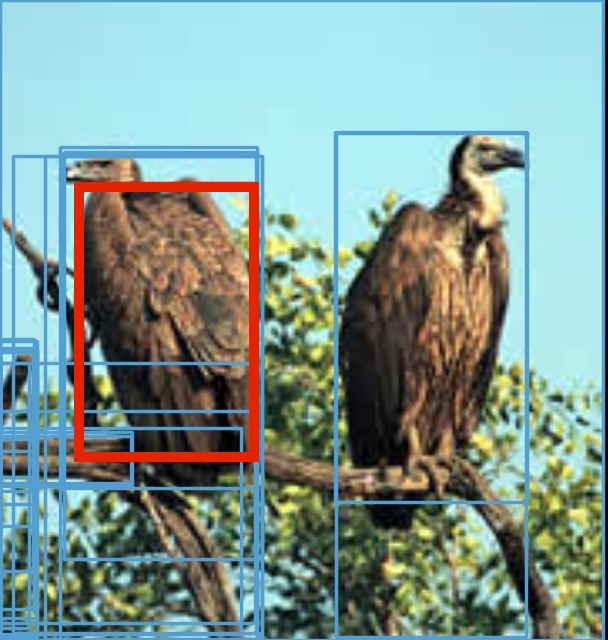} \\
          \rotatebox[origin=c]{90}{\scriptsize Max-size\!\!\!\!\!\!\!\!\!\!\!\!\!\!\!\!\!\!\!\!\!\!\!\!\!\!\!\!\!\!\!\!}
          & \includegraphics[width=0.17\linewidth]{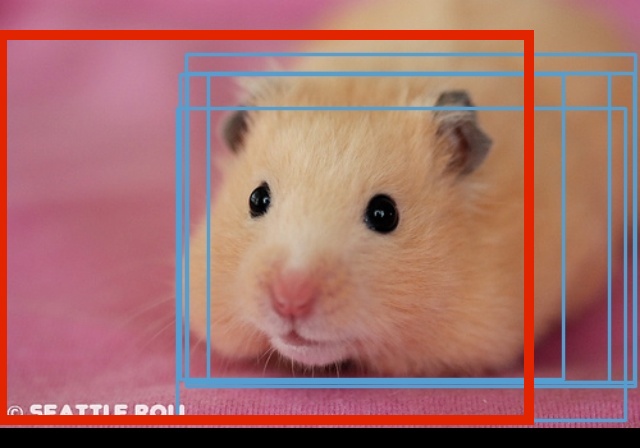}
          & \includegraphics[width=0.17\linewidth]{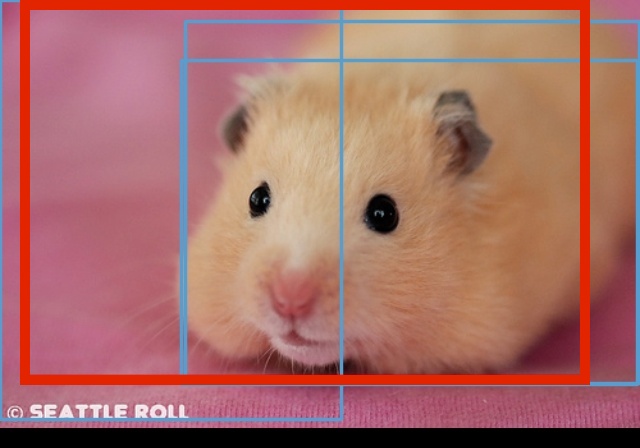}
          & \includegraphics[width=0.17\linewidth]{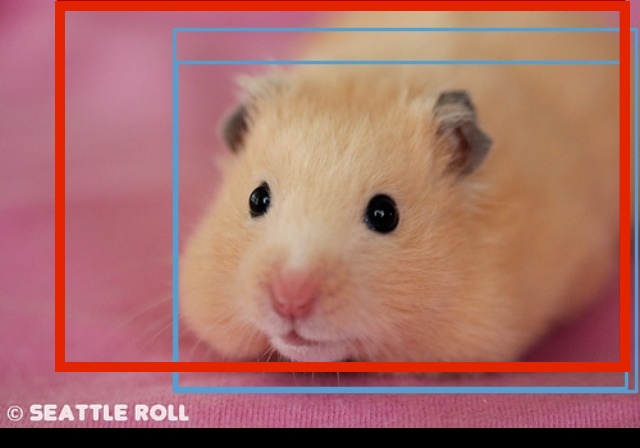}
          & \includegraphics[trim={0 1cm 0 4.0cm}, clip, width=0.15\linewidth]{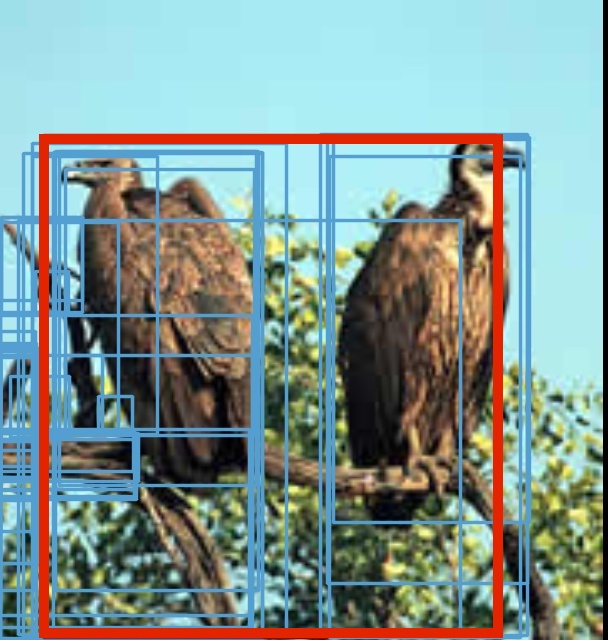}
          & \includegraphics[trim={0 1cm 0 4.0cm}, clip, width=0.15\linewidth]{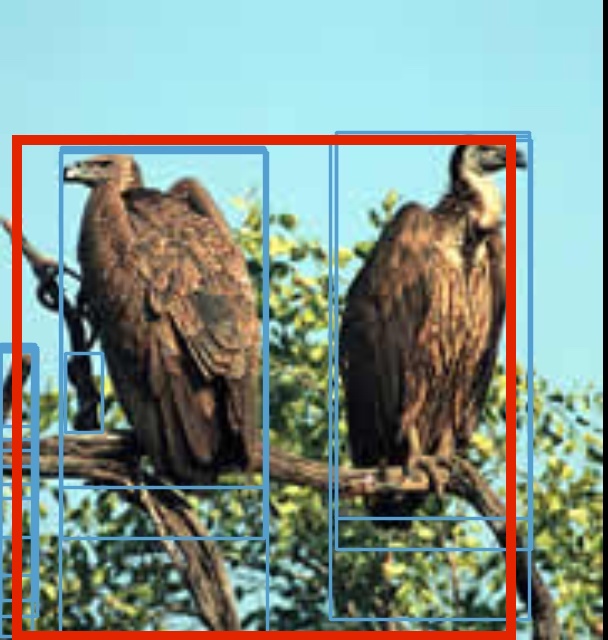}
          & \includegraphics[trim={0 1cm 0 4.0cm}, clip, width=0.15\linewidth]{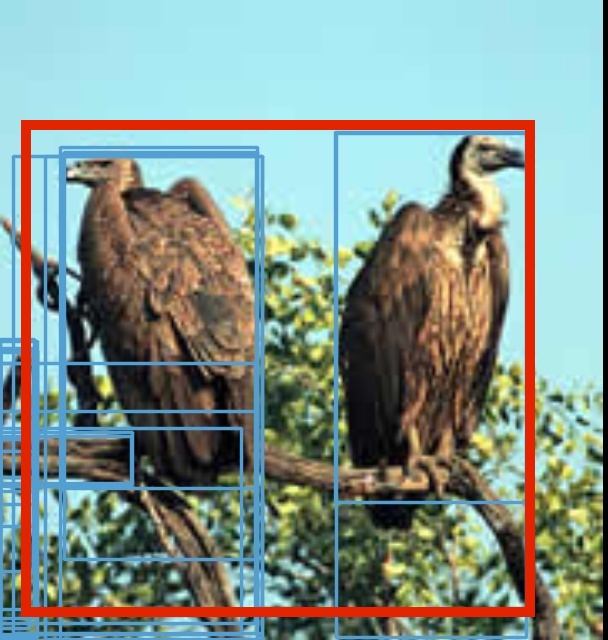} \\
          \end{tabular}
          \vspace{\captionSpace}
          \caption{
            \textbf{Visualization of the assigned boxes during training.} We show all boxes with score $>0.5$ in \textcolor{skyblue3}{blue} and the assigned (selected) box in \textcolor{DarkScarletRed}{red}.
            \textbf{Top}: The prediction-based method selects different boxes across training, and the selected box may not cover the objects in the image.
            \textbf{Bottom}: Our simpler max-size variant selects a box that covers the objects and is more consistent across training.
            }
          \lblfig{qualitative-proposal}
          \vspace{-5mm}
  \end{figure}

  \begin{table}[!b]
    \vspace{-5mm}
    \begin{center}
    \begin{tabular}{@{}l@{\ \ \ \ \ \ \ \ \ }c@{\ \ \ \ \ \ }c@{\ \ \ \ \ \ }c@{\ \ \ \ \ \ }c@{}}
    \toprule
     & mAP$\mask$ & mAP$\mask_{\text{novel}}$ & mAP$\mask_{\text{c}}$ & mAP$\mask_{\text{f}}$ \\
    \midrule
    ViLD-text~\cite{gu2021zero} & 24.9 & 10.1 & 23.9 & \bf 32.5 \\
    ViLD~\cite{gu2021zero} & 22.5 & 16.1 & 20.0 & 28.3 \\
    ViLD-ensemble~\cite{gu2021zero} & 25.5 & 16.6 & 24.6 & 30.3 \\
    \midrule
    \OURS{} & \bf 26.8 & \bf 17.8 & \bf 26.3 & 31.6 \\
    \bottomrule
    \end{tabular}
    \end{center}
    \vspace{\captionSpace}
    \caption{\textbf{\Zeroshot{} \lvis{} compared to \vild~\cite{gu2021zero}.} We train our model \emph{using their training settings and architecture} (MaskRCNN-ResNet50, training from scratch).
    We report mask mAP and its breakdown to novel (rare), common, and frequent classes.
    Variants of \vild{} use distillation (\vild) or ensembling (\vild-ensemble.).
    \OURS{} (with IN-L) uses a single model and improves both mAP and mAP$_{\text{novel}}$.
    }
    \lbltab{main-lvis-vild}
\end{table}

\subsection{Comparison with a fully-supervised detector}
In~\cref{tbl:imagelabel}, compared with the strong baseline \baselineDet{}, \OURS{} improves the detection performance by $2.4$ mAP and $8.3$ mAP$_\text{novel}$.
Thus, \OURS{} with image-level labels leads to strong \zeroshot{} detection performance and can provide orthogonal gains to existing \zeroshot{} detectors~\cite{bansal2018zero}.
To further understand the \zeroshot{} capabilities of \OURS{}, we also
report the \emph{top-line} results trained with box labels for all classes (\cref{tbl:imagelabel} last row). 
Despite not using box labels for the novel classes, \OURS{} with ImageNet
performs favorably compared to the fully-supervised detector.
This result also suggests that bounding box annotations may not be required for new classes.
\OURS{} combined with large image classification datasets is a simple and effective alternative for increasing detector vocabulary.

\subsection{Comparison with the state-of-the-art}

We compare \OURS{}'s \zeroshot{} object detectors with state-of-the-art methods on the \zeroshot{} \lvis and the \zeroshot{} COCO benchmarks.
In each case, we strictly follow the architecture and setup from prior work to ensure fair comparisons.

\par \noindent\textbf{\Zeroshot{} LVIS.}
We compare to \vild~\cite{gu2021zero},
 which first uses CLIP embeddings~\cite{radford2021learning}  for \zeroshot{} detection.
We strictly follow their training setup and model architecture (\supplement{sec:vild-details}) and report results in~\cref{tbl:main-lvis-vild}.
Here \vild-text is exactly our Box-Supervised baseline.
 \OURS{} provides a gain of $7.7$ points on mAP$_\text{novel}$.
Compared to \vild-text, \vild{}, which uses knowledge distillation from the CLIP visual backbone, improves mAP$_\text{novel}$ at the cost of hurting overall mAP.
Ensembling the two models, \vild-ens{} provides improvements for both metrics.
On the other hand, \OURS{} uses a single model which improves both novel and overall mAP, and outperforms the \vild{} ensemble.

\par \noindent\textbf{\Zeroshot{} COCO.}
Next, we compare with prior works on the popular \zeroshot{} COCO benchmark~\cite{bansal2018zero}
(see benchmark and implementation details in \supplement{sec:coco-details}).
We strictly follow OVR-CNN~\cite{zareian2021open} to use Faster R-CNN with ResNet50-C4 backbone and do not use any improvements from~\cref{sec:implementation_details}.
Following~\cite{zareian2021open}, we use COCO captions
as the image-supervised data.
We extract nouns from the captions and use both the image labels and captions as supervision.

\begin{table}[!t]
    \begin{center}
    % \small
    % \vspace{-5mm}
    \begin{tabular}{@{}l@{\ \ \ \ \ \ \ \ \ \ \ \ }c@{\ \ \ \ \ \ \ \ \ \ \ \ }c@{\ \ \ \ \ \ \ \ \ \ \ \ }c@{}}
    \toprule
    & mAP50$\bbox_{\text{all}}$ & mAP50$\bbox_{\text{novel}}$ & mAP50$\bbox_{\text{base}}$  \\
    \midrule
    Base-only\dag & 39.9 & 0 & \bf 49.9  \\
    Base-only (CLIP) & 39.3  & 1.3 & 48.7   \\
    WSDDN \cite{bilen2016weakly}\dag & 24.6 &20.5 & 23.4 \\
    Cap2Det \cite{ye2019cap2det}\dag & 20.1 & 20.3 & 20.1 \\
    SB \cite{bansal2018zero}\ddag & 24.9 & 0.31 & 29.2    \\
    DELO \cite{zhu2020don}\ddag  & 13.0 & 3.41 & 13.8   \\
    PL \cite{rahman2020improved}\ddag & 27.9 & 4.12 & 35.9   \\
    {OVR-CNN}~\cite{zareian2021open}\dag & {39.9}  & {22.8} & {46.0} \\
    \midrule
    \OURS{} & \bf 45.0 & \bf 27.8 & 47.1 \\
    \bottomrule
    \end{tabular}
    \end{center}
    \vspace{\captionSpace}
    \caption{\textbf{\Zeroshot{} COCO~\cite{bansal2018zero}.}
    We compare \OURS{} using the same training data and architecture from OVR-CNN~\cite{zareian2021open}.
    We report box mAP at IoU threshold 0.5 using Faster R-CNN with ResNet50-C4 backbone.
    \OURS{} builds upon the CLIP baseline (second row) and shows significant improvements over prior work.
    \dag: results quoted from OVR-CNN~\cite{zareian2021open} paper or code. \ddag: results quoted from the original publications.
    }
    \lbltab{zs-coco}
    \vspace{-8mm}
\end{table}

~\reftab{zs-coco} summarizes our results.
As the training set contains only 48 base classes, the base-class only model (second row) yields low mAP on novel classes.
\OURS{} improves the baseline and outperforms OVR-CNN~\cite{zareian2021open} by a large margin, using exactly the same model, training recipe, and data.

Additionally, similar to~\cref{tbl:imagelabel}, we compare to prior prediction-based methods on the \zeroshot{} COCO benchmark in \supplement{sec:coco-details}.
In this setting too, \OURS{} improves over prior work providing significant gains on novel class detection and overall detection performance.

\subsection{Detecting 21K classes across datasets without finetuning}
\lblsec{crossdataset}

Next, we train a detector with the full 21K classes of ImageNet.
We use our strong recipe with \swinB{}~\cite{liu2021swin} backbone.
In practice, training a classification layer of 21K classes is computationally involved.\footnote{This is more pronounced in detection than classification, as the ``batch-size'' for the classification layer is $512 \times$ image-batch-size, where $512$ is \#RoIs per image.}
We adopt a modified Federated Loss~\cite{zhou2021probabilistic} that uniformly samples $50$ classes from the vocabulary at every iteration.
We only compute classification scores and back-propagate on the sampled classes.

\begin{table}[!t]
    \begin{center}
    \begin{tabular}{@{}l@{\ \ \ \ \ \ }c@{\ \ \ \ \ }c@{\ \ \ \ \ \ }c@{\ \ \ \ }c@{}}
    \toprule
     & \multicolumn{2}{c}{Objects365~\cite{shao2019objects365}} & \multicolumn{2}{c}{OpenImages~\cite{kuznetsova2020open}} \\
     & mAP$\bbox$ & mAP$\bbox_{\text{rare}}$ & mAP50$\bbox$ & mAP50$\bbox_{\text{rare}}$  \\
     \cmidrule(r){1-1}
     \cmidrule(r){2-3}
     \cmidrule(r){4-5}
    \baselineDet      & 19.1 & 14.0 & 46.2 & 61.7 \\
    \OURS{} w. IN-L   & 21.2 & 17.8 & 53.0 & 67.1 \\
    \OURS{} w. IN-21k & \bf 21.5 & \bf 20.0 & \bf 55.2 & \bf 68.8 \\
    \cmidrule(r){1-1}
    \cmidrule(r){2-3}
    \cmidrule(r){4-5}
    \color{gray} {Dataset-specific oracles} & \color{gray} 31.2 & \color{gray} 22.5 & \color{gray} 69.9 & \color{gray} 81.8 \\
    \bottomrule
     \end{tabular}
    %  }
    \end{center}
    \vspace{\captionSpace}
    \caption{\textbf{Detecting 21K classes across datasets.}
    We use \OURS{} to train a detector and evaluate it on multiple datasets \emph{without retraining}.
    We report the bounding box mAP on Objects365 and OpenImages.
    Compared to the \baselineDet{} baseline (trained on \lvis-all),
    \OURS{} leverages image-level supervision to train robust detectors.
    The performance of \OURS{} is $70\%$-$80\%$ of {\color{gray}dataset-specific models} (bottom row) that use dataset specific box labels.
    }
    \lbltab{crossdataset}
    \vspace{-8mm}
    \end{table}

    \begin{figure*}[!b]
        \vspace{-5mm}
        \centering
              \begin{tabular}{@{}c@{\ }c@{\ }c@{\ }c@{\ }}
              \includegraphics[trim={0 0 0 4.0cm}, clip, width=0.24\textwidth]{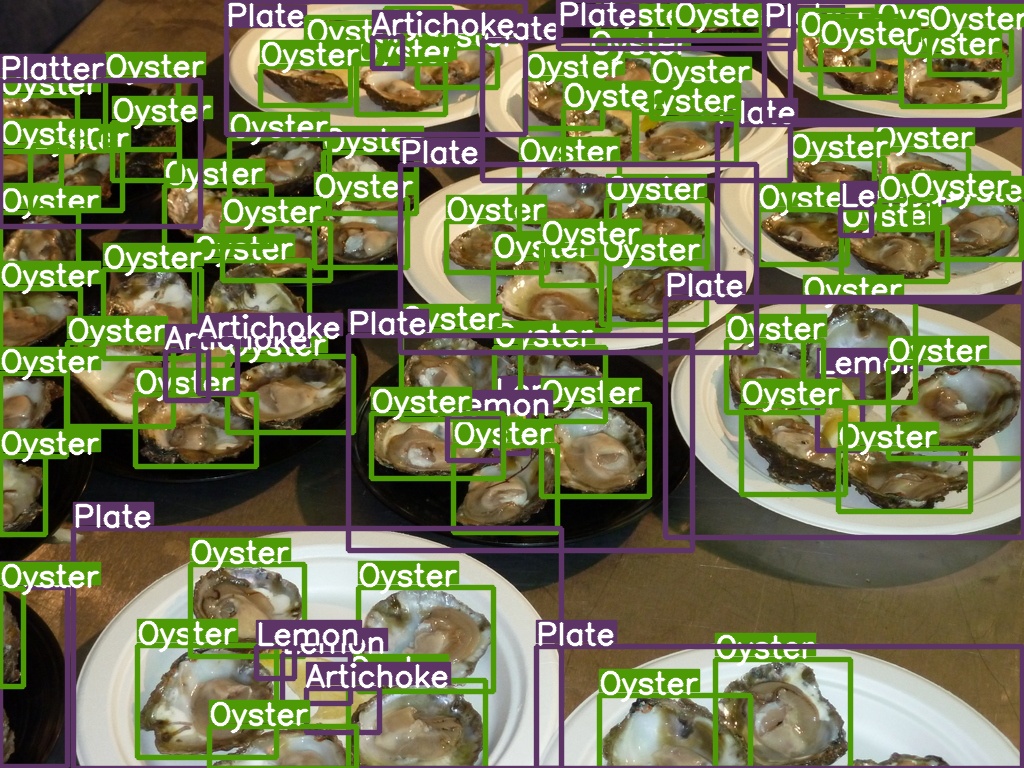} 
              & \includegraphics[trim={0 0 0 4.0cm}, clip, width=0.24\textwidth]{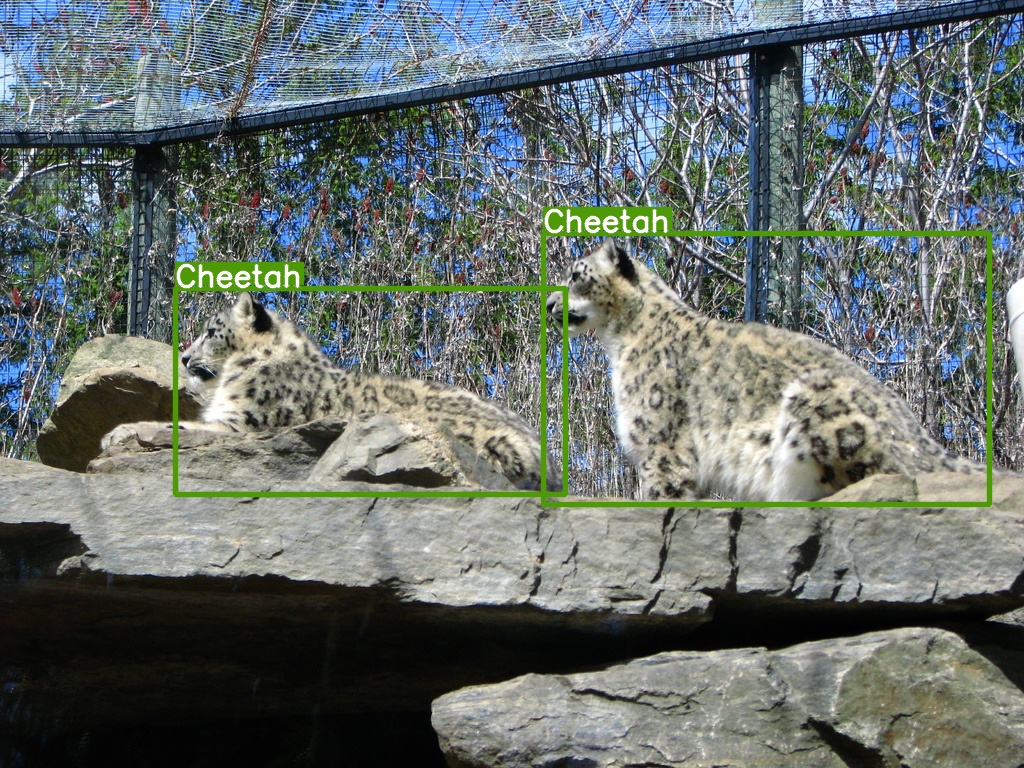}
              & \includegraphics[trim={0 0 0 2.3cm}, clip, width=0.24\textwidth]{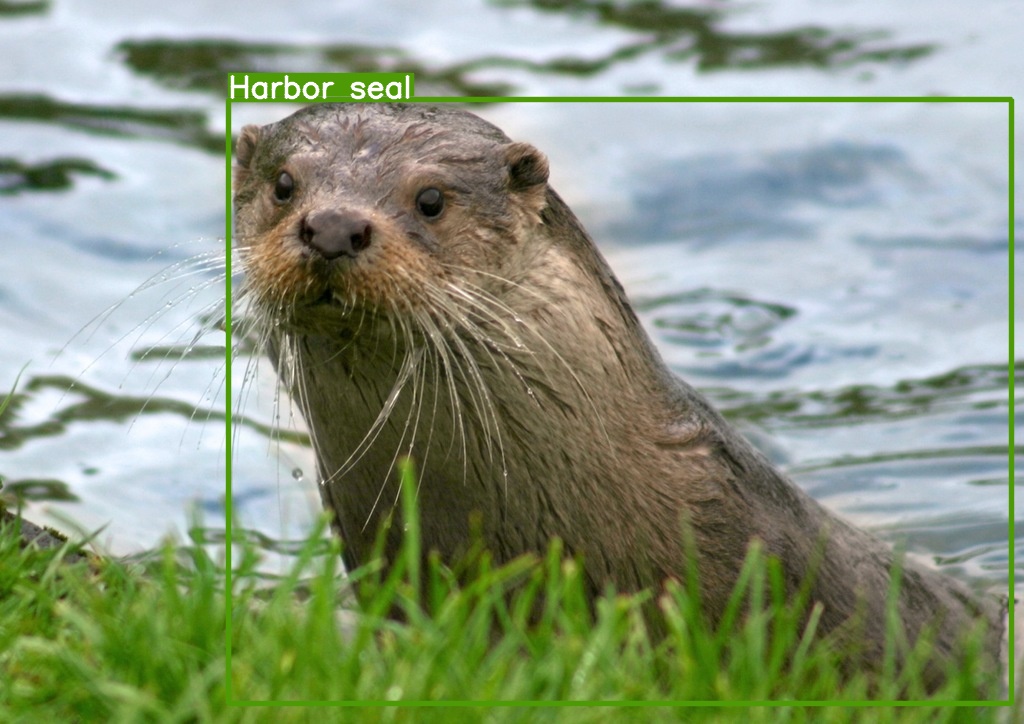}
              &\includegraphics[trim={0 0 0 0.0cm}, clip, width=0.24\textwidth]{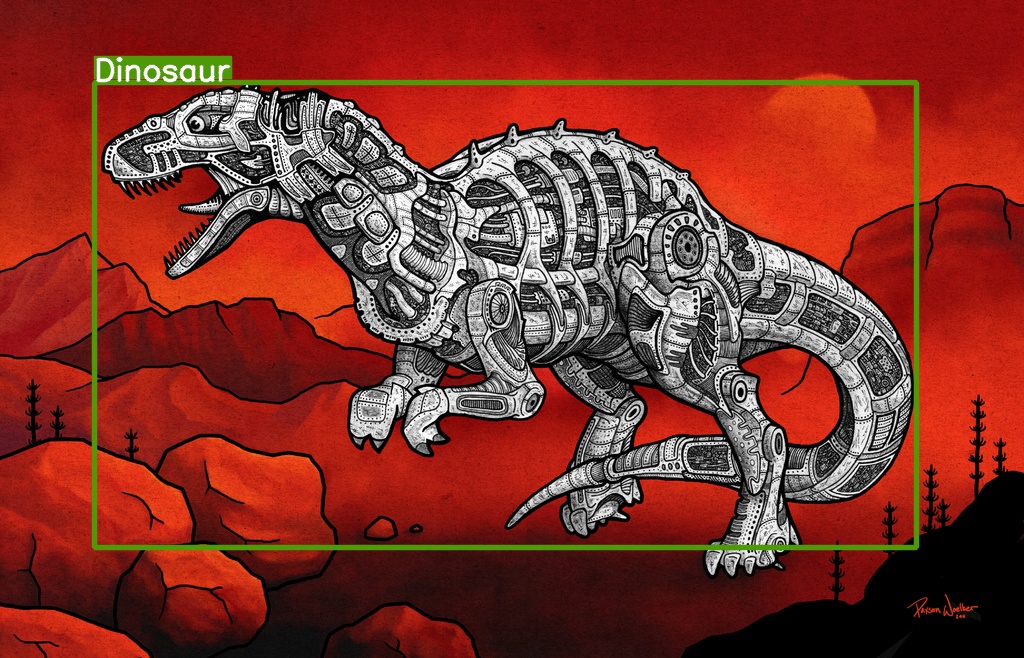} \\
              \includegraphics[width=0.24\textwidth]{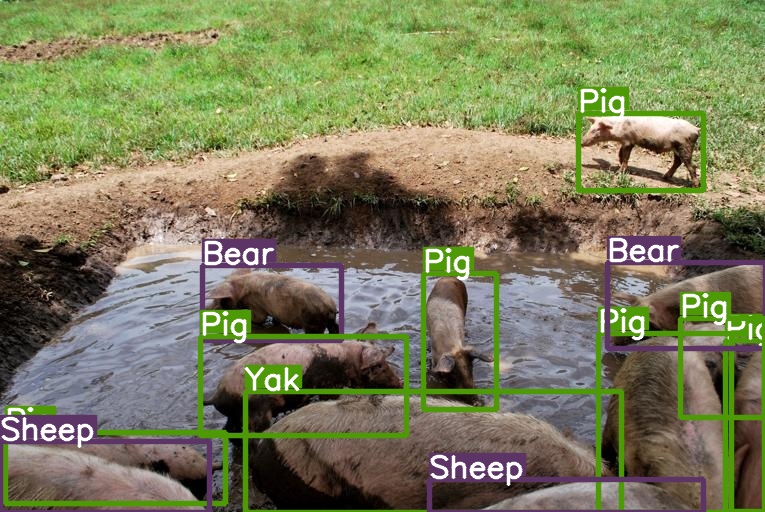}
              &\includegraphics[trim={0 2.0cm 0 0}, clip, width=0.24\textwidth]{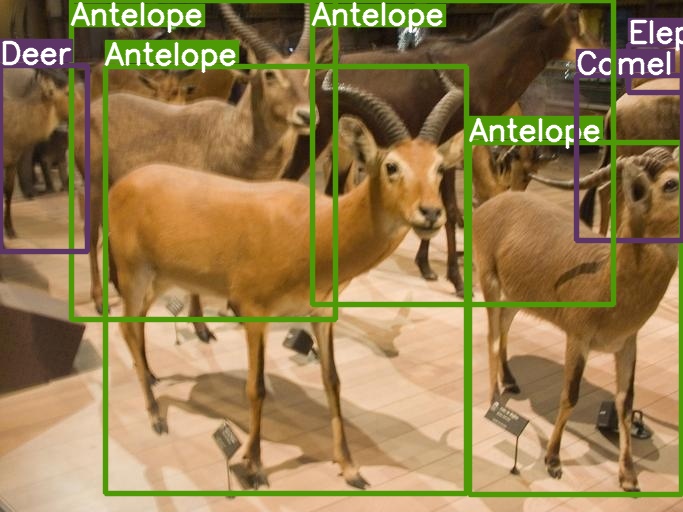}
              &\includegraphics[width=0.24\textwidth]{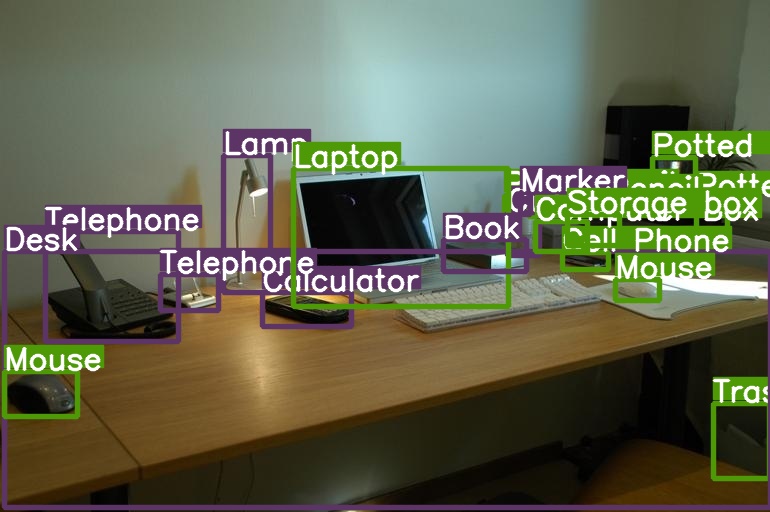}
              &\includegraphics[trim={0 0 0 2.0cm}, clip, width=0.24\textwidth]{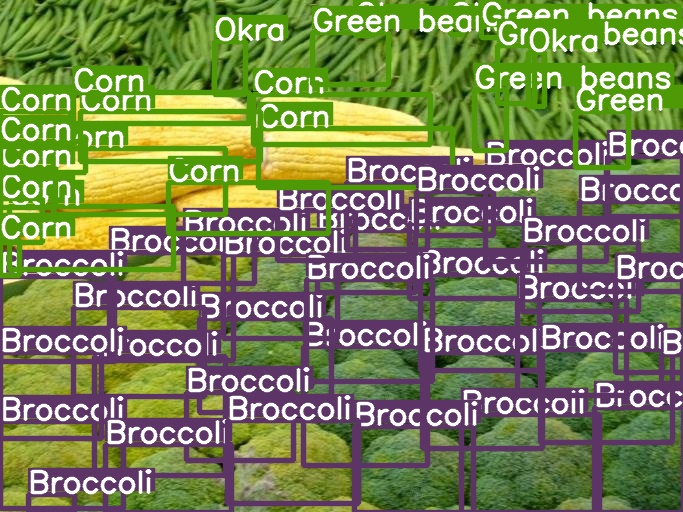}\\
              \end{tabular}
              \vspace{\captionSpace}
              \caption{
                \textbf{Qualitative results of our 21k-class detector.}
              We show random samples from images containing novel classes in OpenImages (top) and Objects365 (bottom) validation sets.
              We use the CLIP embedding of the corresponding vocabularies.
              We show
              LVIS classes
              in \textcolor{plum3}{purple} and novel classes in \textcolor{green4}{green}.
              We use a score threshold of $0.5$ and
              show the most confident class for each box.
              Best viewed on screen.}
              \lblfig{qualitative}
      \end{figure*}

As there are no direct benchmark to evaluate detectors with such large vocabulary,
we evaluate our detectors on new datasets \emph{without finetuning}.
We evaluate on two large-scale object detection datasets:
Objects365v2~\cite{shao2019objects365} and OpenImages~\cite{kuznetsova2020open}, both with around $1.8$M training images.
We follow LVIS to split $\frac{1}{3}$ of classes with the fewest training images as rare classes.
\reftab{crossdataset} shows the results.
On both datasets, \OURS{} improves the
Box-Supervised baseline by a large margin, especially on classes with fewer annotations.
Using all the 21k classes further improves performance owing to the large vocabulary.
Our single model significantly reduces the gap towards the dataset-specific oracles
and reaches $70\%$-$80\%$ of their performance without using the corresponding $1.8$M detection annotations.
See ~\reffig{qualitative} for qualitative results.

\subsection{Ablation studies}
\label{sec:ablation}

We now ablate our key components under the \zeroshot{} LVIS setting with IN-L as the image-classification data.
We use our strong training recipe as described in \cref{sec:implementation_details} for all these experiments.

\begin{table}[!t]
\begin{center}
\begin{tabular}{@{}l@{\ \ \ \ \ \ \ \ }c@{\ \ \ \ \ \ \ \ }c@{\ \ \ \ \ \ \ \ }c@{\ \ \ \ \ \ \ \ }c@{}}
\toprule
& \multicolumn{2}{c}{Box-Supervised} & \multicolumn{2}{c}{\OURS{}} \\
 Classifier & mAP$\mask$ & mAP$\mask_{\text{novel}}$  & mAP$\mask$ & mAP$\mask_{\text{novel}}$ \\
\cmidrule(r){1-1}
\cmidrule(r){2-3}
\cmidrule(r){4-5}
 *CLIP~\cite{radford2021learning}  & 30.2 & 16.4 & 32.4 & 24.9 \\
 Trained & 27.4 & 0 & 31.7  & 17.4 \\
 FastText~\cite{joulin2016fasttext} & 27.5 & 9.0 & 30.9 & 19.2 \\
 OpenCLIP~\cite{openclip} & 27.1 & 8.9  & 30.7 & 19.4  \\
\bottomrule
 \end{tabular}
\end{center}
\vspace{\captionSpace}
\caption{
  \textbf{\OURS{} with different classifiers.}
  We vary the classifier used with \OURS{} and observe that it works well with different choices.
  While CLIP embeddings give the best performance (* indicates our default),
  all classifiers benefit from our \OURS{}.
}
\lbltab{classifier}
\vspace{-8mm}
\end{table}

\vspace{0.03in}
\par \noindent \textbf{Classifier weights.}
We study the effect of different classifier weights $\bW$.
While our main \zeroshot{} experiments use CLIP~\cite{radford2021learning},
we show the gain of \OURS{} is independent of CLIP.
We train \baselineDet{} and \OURS{} with different classifiers, including a standard random initialized and trained classifier, and other \emph{fixed} language models~\cite{joulin2016fasttext,openclip}
The results are shown in~\reftab{classifier}.
By default, a trained classifier cannot recognize novel classes.
However, \OURS{} enables novel class recognition ability even in this setting ($17.4$ mAP$_{\text{novel}}$ for classes without detection labels).
Using language models such as FastText~\cite{joulin2016fasttext}
or an open-source version of CLIP~\cite{openclip}
leads to better novel class performance.
CLIP~\cite{radford2021learning} performs the best among them.

\begin{table}[!b]
  \vspace{-5mm}
\begin{center}
\begin{tabular}{@{}l@{\ \ \ \ \ \ \ \ \ }c@{\ \ \ \ \ \ \ \ \ \ \ \ }c@{\ \ \ \ \ \ \ \ \ \ \ \ }c@{}}
\toprule
 & Pretrain data & mAP$\mask$ & mAP$\mask_{\text{novel}}$ \\
\midrule
\baselineDet & IN-1K & 26.1 & 13.6 \\
\OURS{} & IN-1K & 28.8 \scriptsize \textcolor{green4}{(+2.7)} &
 21.7 \scriptsize \textcolor{green4}{(+8.1)}\\
\midrule
\baselineDet & IN-21K & 30.2 & 16.4 \\
\OURS{} & IN-21K & 32.4 \scriptsize \textcolor{green4}{(+2.2)} & 24.9 \scriptsize \textcolor{green4}{(+8.5)} \\
\bottomrule
 \end{tabular}
\end{center}
\vspace{\captionSpace}
\caption{
  \textbf{\OURS{} with different pretraining data.}
  Top: our method using ImageNet-1K as pretraining and ImageNet-21K as co-training;
  Bottom: using ImageNet-21K for both pretraining and co-training. Co-training helps pretraining in both cases. }
\lbltab{pretrain}
\end{table}

\vspace{0.03in}
\par \noindent \textbf{Effect of Pretraining.}
Many existing methods use additional data only for pretraining ~\cite{zhang2021mosaicos,zareian2021open,desai2021virtex},
while we use image-labeled data for co-training.
We present results of \OURS{} with different types of pretraining in~\reftab{pretrain}.
\OURS{} provides similar gains across different types of pretraining, suggesting that our gains are orthogonal to advances in pretraining.
We believe that this is because pretraining improves the overall features, while \OURS{} uses co-training which improves both the features and the classifier.

\subsection{The standard \lvis{} benchmark}
Finally, we evaluate \OURS{} on the standard \lvis{} benchmark~\cite{gupta2019lvis}.
In this setting, the baseline (\baselineDet) is trained with box and mask labels for all classes while \OURS{} uses additional image-level labels from \imnetLvis{}.
We train Detic with the same recipe in ~\cref{sec:implementation_details} and use a strong Swin-B~\cite{liu2021swin} backbone and $896\times896$ input size.
We report the mask mAP across all classes and also split into rare, common, and frequent classes.
Notably, \OURS{} achieves $41.7$ mAP and $41.7$ \mAPr, closing the gap between the overall mAP and the rare mAP.
This suggests \OURS{} effectively uses image-level labels to improve the performance of classes with very few boxes labels.
\supplement{sec:full-lvis} provides more comparisons to prior work~\cite{zhang2021mosaicos} on LVIS.
\supplement{sec:detr} shows \OURS{} generalizes to DETR-based~\cite{zhu2020deformable} detectors.

\begin{table}[!t]
    \begin{center}
    \begin{tabular}{@{}l@{\ \ }c@{\ \ \ }c@{\ \ \ }c@{\ \ \ }c@{\ \ \ }c@{}}
    \toprule
     & Backbone& mAP$\mask$ & mAP$\mask_{\text{r}}$ & mAP$\mask_{\text{c}}$ & mAP$\mask_{\text{f}}$\\
    \midrule
    MosaicOS$\dagger$~\cite{zhang2021mosaicos} & ResNeXt-101 & 28.3 & 21.7 & 27.3 & 32.4 \\
    CenterNet2~\cite{zhou2021probabilistic} & ResNeXt-101 & 34.9 & 24.6 & 34.7 & 42.5\\
    AsyncSLL$\dagger$~\cite{han2020joint} & ResNeSt-269 &36.0 & 27.8 & 36.7 & 39.6 \\
    SeesawLoss~\cite{wang2021seesaw}\!  & ResNeSt-200 & 37.3 & 26.4 & 36.3 & \bf 43.1\\
    Copy-paste~\cite{ghiasi2021simple} & EfficientNet-B7\!\!\!\! & 38.1 & 32.1 & 37.1 & 41.9\\
    Tan et al.~\cite{tan20201st} & ResNeSt-269 & 38.8 & 28.5 & 39.5 & 42.7\\
    \midrule
    Baseline  & \swinB & 40.7 & 35.9 & 40.5 & \bf 43.1 \\
    \OURS{}$\dagger$ & \swinB  & \bf{41.7} & \bf{41.7} & \bf 40.8 & 42.6\\
    \bottomrule
     \end{tabular}
    \end{center}
    \vspace{\captionSpace}
    \caption{\textbf{Standard \lvis{}.}
    We evaluate our baseline (\baselineDet) and \OURS{} using different backbones on the \lvis{} dataset.
    We report the mask mAP.
    We also report prior work on LVIS using large backbone networks (single-scale testing) for references (not for apple-to-apple comparison).
    $\dagger$: detectors using additional data.
    \OURS{} improves over the baseline with increased gains for the rare classes.
    }
    \lbltab{fulllvis}
    \vspace{-8mm}
\end{table}

\section{Limitations and Conclusions}
We present \OURS{} which is a simple way to use image supervision in large-vocabulary object detection.
While \OURS{} is simpler than prior assignment-based weakly-supervised detection methods, it supervises all image labels to the same region and does not consider overall dataset statistics.
We leave incorporating such information for future work.
Moreover, open vocabulary generalization has no guarantees on extreme domains.
Our experiments show \OURS{} improves large-vocabulary detection with various weak data sources, classifiers, detector architectures, and training recipes.
{\small
\par \noindent \textbf{Acknowledgments.}
We thank Bowen Cheng and Ross Girshick for helpful discussions and feedback.
This material is in part based upon work supported by the National Science Foundation under Grant No. IIS-1845485 and IIS-2006820.
Xingyi is supported by a Facebook PhD Fellowship.
}

{
\small
\bibliographystyle{splncs04}
\bibliography{egbib}
}

\clearpage

\appendix

\section{Region proposal quality}
\lblsec{proposal}

\begin{table}[!t]
    \small
    \begin{center}
    \begin{subtable}[t]{\linewidth}
    \begin{center}
    \begin{tabular}{@{}l@{}c@{\ }c@{\ \ }c@{\ \ }c@{\ \ }c@{}}
    \toprule 
     & AR$_{r}$50@100 & AR$_{r}$50@300 & AR$_{r}$50@1k & AR50@1k \\
    \midrule
    LVIS-all & 63.3 & 76.3 & 79.7 & 80.9 \\
    LVIS-base & 62.2 & 76.2 & 78.5 & 81.0\\
    \bottomrule
     \end{tabular}
    \caption{
    \textbf{Proposal networks trained with (top) and without (bottom) rare classes.} We report recalls on rare classes and all classes at IoU threshold 0.5 with different number of proposals. Proposal networks trained \emph{without} rare classes can generalize to rare classes in testing.
    }
    \lbltbl{proposal-rare}
    \end{center}
    \end{subtable}
    \begin{subtable}[t]{\linewidth}
        \begin{center}
        \begin{tabular}{@{}l@{\ \ }c@{\ \ \ \ }c@{}}
        \toprule 
         & AR$_{\text{half-1st}}$50@1k & AR$_{\text{half-2nd}}$50@1k \\
        \midrule
        LVIS-half-1st & 80.8 & 69.6 \\ 
        LVIS-half-2nd & 62.9 & 82.2 \\
        \bottomrule
         \end{tabular}
        \caption{
        \textbf{Proposal networks trained on half of the LVIS classes.} We report recalls at IoU threshold 0.5 on the other half classes. Proposal networks produce non-trivial recalls on novel classes.
        }
        \lbltab{proposal-strict}
        \end{center}
        \end{subtable}
    \vspace{\captionSpace}
    \caption{\textbf{Proposal network generalization ability evaluation.} 
    \textbf{(a)}: Generalize from $866$ LVIS base classes to the $337$ rare classes; 
    \textbf{(b)}: Generalize from uniformly sampled half LVIS classes (601/ 602 classes) to the other half.}
    \lbltab{proposal}
    \end{center}
    \vspace{-8mm}
\end{table}

In this section, we show the region proposal network trained on LVIS~\cite{gupta2019lvis}
is satisfactory and generalizes well to new classes by default.
We experiment under our strong baseline in ~\cref{sec:implementation_details}.
\reftab{proposal-rare} shows the proposal recalls with or without rare classes in training.
First, we observe the recall gaps between the two models on rare classes are small (79.7 vs. 78.5);
second, the gaps between rare classes and all classes are small (79.7 vs. 80.9);
third, the absolute recall is relatively high 
($\sim\!80\%$, note recall at IoU threshold 0.5 can be translated into oracle 
mAP-pool~\cite{dave2021evaluating} given perfect classifier and regressor).
All observations indicate the proposals have good generalization abilities to new classes even though they are supervised to background during training.
We consider the proposal generalization is currently not the performance bottleneck in \zeroshot{} detection.
This especially the case as modern detectors use an over-sufficient number of proposals in testing (1K proposals for $<$ 20 objects per image). 
Our observations are consistent with ViLD~\cite{gu2021zero}.

We in addition evaluate a more strict setting, where we uniformly split LVIS classes into two halves. 
I.e., we use classes ID $1, 3, 5, \cdots$ as the first half, and the rest as the second half. 
These two subsets have completely different definitions of ``objects''.
We then train a proposal network on each of them, and evaluate on both subsets.
As shown in \reftab{proposal-strict}, the proposal networks give non-trivial recalls at the complementary other half ($69.6\%$ over $82.2\%$ percent of the full generalizability).
This again supports proposal networks trained on a diverse vocabulary learned a general concept of objects.

\section{Direct captions supervision}
\lblsec{caption}

\begin{table}[!t]
\begin{center}
\begin{tabular}{@{}l@{\ \ \ \ }l@{\ }c@{}c@{}}
\toprule 
 & Supervision & mAP$^\text{mask}$ & mAP$^\text{mask}_{\text{novel}}$ \\
\midrule
Box-Supervised & - & 30.2 & 16.4 \\
\OURS{} w. CC & Image label & \bf 31.0 & 19.8 \\
\OURS{} w. CC & Caption & 30.4 & 17.4 \\
\OURS{} w. CC & Both & \bf 31.0 & \bf 21.3 \\
\midrule
 &   & mAP50$^{\text{box}}_{\text{all}}$ & mAP50$^{\text{box}}_{\text{novel}}$ \\
 \midrule
Box-Supervised & - & 39.3 & 1.3 \\
\OURS{} w. COCO-cap. & Image label & 44.7 & 24.1   \\
\OURS{} w. COCO-cap. & Caption & 43.8  & 21.0  \\
\OURS{} w. COCO-cap. & Both & \bf 45.0  & \bf 27.8 \\
\bottomrule
 \end{tabular}
\end{center}
\vspace{\captionSpace}
\caption{\textbf{Direct caption supervision.} Top: Open-vocabulary LVIS with Conceptual Caption as weakly-labeled data; Bottom block: Open-vocabulary COCO with COCO-caption as weakly-labeled data. Directly using caption embeddings as a classifier is helpful on both benchmarks; the improvements are complementary to \OURS{}.}
\lbltab{caption}
\vspace{-8mm}
\end{table}

As we are using a language model CLIP~\cite{radford2021learning} as the classifier,
our framework can seamlessly incorporate the free-form caption text as image-supervision.
Using the notations in ~\cref{sec:image}, here 
$\dWeak = \{(\bI, t)_i\}$ where $t$ is a free-form text.
In our open-vocabulary detection formulation, text $t$ can natrually be converted to an embedding by the CLIP~\cite{radford2021learning} language encoder $\mathcal{L}$: $w = \mathcal{L}(t)$.
Given a minibatch of $B$ samples $\{(\bI, t)_i\}_{i=1}^{B}$, we compose a dynamic classification layer by stacking all caption features within the batch
$\widetilde{\bW} = \mathcal{L}(\{t_i\}_{i=1}^{B})$.
For the $i$-th image in the minibatch, its ``classification'' label is the $i$-th text, and other texts are negative samples.
We use the injected whole image box to extract RoI feature $\bof'_i$ for image $i$.
We use the same binary cross entropy loss as classifying image labels:
$$L_{cap} = \sum_{i=1}^{B} BCE(\widetilde{\bW} \bof'_{i}, i) $$
We do not back-propagate into the language encoder.

We evaluate the effectiveness of the caption loss in \reftab{caption} on both \zeroshot{} LVIS and COCO (see dataset details in \supplement{sec:coco-details}). 
We compare individually applying the max-size loss for image labels and the caption loss, and applying both of them.
Both image labels and captions can improve both overall mAP and novel class mAP.
Combining both losses gives a more significant improvement. 
Our \zeroshot{} COCO results in \reftab{zs-coco} uses both the max-size loss and the caption loss.

\section{LVIS baseline details}
\lblsec{lvis-baseline}
\begin{table}[!t]
\small
\begin{center}
\begin{tabular}{@{}l@{}c@{}c@{}c@{}c@{}c@{}}
\toprule 
 & mAP$^{\text{box}}$ & mAP$^{\text{box}}_{\text{r}}$ & mAP$^{\text{mask}}$  & mAP$^{\text{mask}}_{\text{r}}$ & T \\
\midrule
D2 baseline~\cite{wu2019detectron2} & 22.9 & 11.3 & 22.4 & 11.6 & 12h \\
+Class-agnostic box\&mask & 22.3 & 10.1 & 21.2 & 10.1 & 12h \\
+Federated loss~\cite{zhou2021probabilistic} & 27.0 & 20.2 & 24.6 & 18.2 & 12h \\
+CenterNet2~\cite{zhou2021probabilistic} & 30.7 & 22.9 & 26.8 & 19.4 & 13h \\
+LSJ $640\!\times\!640$, $4\!\times\!$ sched.\!~\cite{ghiasi2021simple}\!\!\! & 31.0 & 21.6 & 27.2 & 20.1 & 17h \\
+CLIP classifier~\cite{radford2021learning} & 31.5 & 24.2 & 28 & 22.5 & 17h\\
+Adam optimizer, lr$2e$-$4$~\cite{kingma2014adam} & 30.4 & 23.6 & 26.9 & 21.4 & 17h \\
\rowcolor{aluminium1}+IN-21k pretrain~\cite{ridnik2021imagenet21k}* & 35.3 & 28.2 & 31.5 & 25.6 & 17h\\
\color{gray} +Input size $896\!\times\!896$ & \color{gray} 37.1 & \color{gray} 29.5 
& \color{gray} 33.2 & \color{gray} 26.9 & \color{gray} 25h\\
\color{gray} +Swin-B backbone~\cite{liu2021swin} & \color{gray} 45.4 & 
\color{gray} 39.9 & \color{gray} 40.7 & \color{gray} 35.9 & \color{gray} 43h\\
\midrule
\rowcolor{aluminium1}*Remove rare class ann.\cite{gu2021zero}& 33.8 & 17.6 & 30.2 & 16.4 & 17h\\
\bottomrule
 \end{tabular}
\end{center}
\vspace{\captionSpace}
\caption{
\textbf{LVIS baseline evolution.}
First row: the configuration from the detectron2 model zoo.
The following rows change components one by one. 
Last row: removing rare classes from the ``+IN-21k pretrain*'' row.
The two \ctext[RGB]{238,238,236}{gray-filled rows} are the baselines in our main paper, for full LVIS and open-vocabulary LVIS, respectively.
We show rough wall-clock training times ($T$) on our machine with 8 V100 GPUs in the last column.}
\lbltab{lvis-baseline}
\vspace{-8mm}
\end{table}

We first describe the standard LVIS baseline from 
the detectron2 model zoo\footnote{\url{https://github.com/facebookresearch/detectron2/blob/main/configs/LVISv1-InstanceSegmentation/mask_rcnn_R_50_FPN_1x.yaml}}.
% , which is exactly the baseline used in MosaicOS~\cite{zhang2021mosaicos}.
This baseline uses ResNet-50 FPN backbone and a $2\times$ training schedule 
($180k$ iterations with batch-size $16$)\footnote{We are aware different projects use different notations of a $1\times$ schedule. In this paper we always refer $1\times$ schedule to $16 \times 90k$ images}.
Data augmentation includes horizontal flip and random resize short side [$640$, $800$], long side $<1333$.
The baseline uses SGD optimizer with a learning rate 0.02 (dropped by $10\!\times$ at $120k$ and $160k$ iteration).
The bounding box regression head and the mask head are class-specific.

~\reftab{lvis-baseline} shows the roadmap from the detectron2 baseline to 
our baseline (\cref{sec:implementation_details}).
First, we prepare the model for new classes by making the box and mask heads class-agnostic.
This slightly hurts performance.
We then use Federated loss~\cite{zhou2021probabilistic}
and upgrade the detector to CenterNet2~\cite{zhou2021probabilistic} 
(i.e., replacing RPN with CenterNet and multiplying proposal score to classification score).
Both modifications improve mAP and \mAPr significantly, 
and CenterNet2 slightly increases the training time.

Next, we use the EfficientDet~\cite{tan2020efficientdet,ghiasi2021simple} 
style large-scale jittering and train a longer schedule ($4\times$). 
To balance the training time, we also reduce the training image size to 
$640 \times 640$ (the testing size is unchanged at $800\times1333$) 
and increase batch-size to $64$ (with the learning rate scaled up to $0.08$).
The resulting augmentation and schedule is slightly better than the default multi-scale training,
with $30\%$ more training time.
A longer schedule is beneficial when using more data, and can be improved by 
larger resolution.

Next, we switch in the CLIP classifier~\cite{radford2021learning}. We follow ViLD~\cite{gu2021zero} to L2 normalize the embedding and RoI feature before dot-product.
Note CenterNet2 uses a cascade classifier~\cite{cai2018cascade}.
We use CLIP for all of them.
Using CLIP classifier improves rare class mAP.

Finally, we use an ImageNet-21k pretrained ResNet-50 model from Ridnik 
\etal~\cite{ridnik2021imagenet21k}.
We remark the ImageNet-21k pretrained model requires using Adam optimizer 
(with learning rate $2e$-$4$).
Combing all the improvements results in $35.3$ mAP$^{\text{box}}$ 
and $31.5$ mAP$^{\text{mask}}$, and trains in a favorable time ($17$h on 8 V100 GPUs).
We use this model as our baseline in the main paper.

Increasing the training resolution or using a larger backbone~\cite{liu2021swin}
can further increase performance significantly, at a cost of longer training time. 
We use the large models only when compared to the state-of-the-art models.

\section{Resolution change for classification data}
\lblsec{ratio-and-size}

\begin{table}[!t]
    \begin{center}
    \begin{tabular}{@{}l@{\ \ }c@{\ \ \ }c@{\ \ \ }c@{\ \ \ }c@{}c@{}}
    \toprule 
     & Ratio & Size & mAP$^\text{mask}$ & mAP$^\text{mask}_{\text{novel}}$ \\
    \midrule
    Bos-Supervised & 1: 0 & - & 30.2 & 16.4 \\
    \midrule
    \OURS{} w. IN-L & 1: 1 & 640 & 30.9 & 23.3 \\
    \OURS{} w. IN-L & 1: 1 & 320 & 32.0 & 24.0\\
    \OURS{} w. IN-L & 1: 4 & 640 & 31.1 & 23.5\\
    \OURS{} w. IN-L & 1: 4 & 320 & \bf 32.4 & \bf 24.9 \\
    \midrule
    \OURS{} w. CC & 1: 1 & 640 & 30.8 & 21.6 \\
    \OURS{} w. CC & 1: 1 & 320 & 30.8 & 21.5 \\
    \OURS{} w. CC & 1: 4 & 640 & 30.7 & 21.0 \\
    \OURS{} w. CC & 1: 4 & 320 & \bf 31.1 & \bf 21.8 \\
    \bottomrule
     \end{tabular}
    \end{center}
    \vspace{\captionSpace}
    \caption{\textbf{Ablations of the resolution change.} 
    We report mask mAP on the open-vocabulary LVIS following the setting of ~\reftab{imagelabel}.
    Top: ImageNet as the image-labeled data.
    Bottom: CC as the image-labeled data.}
    \lbltab{ratio-and-size}
    \vspace{-8mm}
    \end{table}

~\reftab{ratio-and-size} ablates the resolution change in ~\cref{sec:implementation_details}.
Using a smaller input resolution improves $\sim\!1$ point for both mAP and \mAPnoval{} with ImageNet,
but does not impact much with CC.
Using more batches for the weak datasets is slightly better than a $1:1$ ratio.

\section{Prediction-based losses implementation details}
\lblsec{predictionbaseddetails}
Following the notations in ~\cref{sec:image}, we implement the prediction-based weakly-supervised detection losses as below:

\par \noindent \textbf{WSDDN}~\cite{bilen2016weakly} 
learns a soft weight on the proposals to weight-sum the proposal classification scores into a single image classification score:
$$L_{\text{WSDDN}} = BCE(\sum_j (\text{softmax}(\bW'\bF)_j * \bS_j), c)$$
where $\bW'$ is a learnable network parameter.

\par \noindent \textbf{Predicted}~\cite{redmon2017yolo9000} 
selects the proposal with the max predicted score on class $c$:
$$L_{\text{Predicted}} = BCE(\bS_j, c), j = \text{argmax}_j \bS_{jc} $$
\par \noindent \textbf{DLWL*}~\cite{ramanathan2020dlwl} first runs a clustering algorithm with IoU threshold 0.5.
Let $\mathcal{J}$ be the set of peaks of each cluster 
(i.e., the proposal within the cluster and has the max predicted score on class $c$),
We then select the top $N_c=3$ peaks with the highest prediction scores on class $c$.
\begin{align*}
L_{\text{DLWL*}} = \frac{1}{N_c}\sum_{t=1}^{N_c} & BCE(\bS_{j_t}, c), \\
& j_t = \text{argmax}_{j \in \mathcal{J}, j \neq \{j_1, \dots, j_{t - 1}\}} \bS_{jc}\end{align*}
The original DLWL~\cite{ramanathan2020dlwl} in addition upgrades $\bS$ using an IoU-based assignment matrix from self-training and bootstrapping (See their Section 3.2). 
In our implementation, we did not include this part, as our goal is to only compare the training losses.

\section{More comparison between prediction-based and non-prediction-based methods}
\lblsec{comparison-prediction-based}

\begin{table}[!t]
    \begin{center}
    \begin{subtable}[t]{\linewidth}
    \begin{center}
        % \centering
    \begin{tabular}{@{}l@{\ \ }c@{\ }c@{\ }c@{\ }c@{}}
    \toprule 
     & Dataset & Backbone & mAP$^\text{mask}$ & mAP$^\text{mask}_{\text{novel}}$ \\
    \midrule
    Box-Supervised & & & 30.2 & 16.4 \\
    Predicted & \multirow{1}{*}{LVIS-base} & \multirow{1}{*}{Res50} & 31.2 & 20.4 \\
    Max-size &&& 32.4 \scriptsize \textcolor{green4}{(+1.2)} & 24.6 \scriptsize \textcolor{green4}{(+4.2)}\\
    \midrule
    Box-Supervised &  & & 38.4 & 21.9 \\
    Predicted & \multirow{1}{*}{LVIS-base} & \multirow{1}{*}{SwinB} & 40.0 & 31.7 \\
    Max-size & & & 40.7 \scriptsize \textcolor{green4}{(+0.7)} & 33.8 \scriptsize \textcolor{green4}{(+2.1)}\\
    \midrule
    \midrule
    Box-Supervised &  & & 31.5 & 25.6 \\
    Predicted & \multirow{1}{*}{LVIS-all} & \multirow{1}{*}{Res50} & 32.5 & 28.4 \\
    Max-size & & & 33.2 \scriptsize \textcolor{green4}{(+0.7)} & 29.7 \scriptsize \textcolor{green4}{(+1.3)}\\
    \midrule
    Box-Supervised & & & 40.7 & 35.9 \\
    Predicted & \multirow{1}{*}{LVIS-all} & \multirow{1}{*}{SwinB} & 40.6 & 39.8 \\
    Max-size & & & 41.3 \scriptsize \textcolor{green4}{(+0.7)} & 40.9 \scriptsize \textcolor{green4}{(+1.1)}\\
    \bottomrule
     \end{tabular}
    \end{center}
    \vspace{\captionSpace}
    \caption{\textbf{Predicted loss and max-size loss with different prediction qualities.}
    We show the mask mAP of the box-supervised baseline, Predicted loss~\cite{redmon2017yolo9000}, and our max-size loss.
    We show the delta between max-size loss and predicted loss in \textcolor{green4}{green}.
    Improving the backbone and including rare classes in training can both narrow the gap. Max-size consistently performs better.
    }
    \lbltab{max-score-gap}
    \vspace{-3mm}
    \end{subtable}
    \begin{subtable}[t]{\linewidth}
    \begin{center}
        % \centering
    \begin{tabular}{@{}l@{\ \ \ \ }c@{\ \ \ \ }c@{\ \ \ \ }c@{\ \ \ \ }c@{\ \ \ \ }c@{}}
    \toprule
     & \multicolumn{2}{c}{Cover rate} & \multicolumn{3}{c}{Consistency} \\
     & IN-L & COCO & IN-L & CC & COCO\\
     \cmidrule(r){1-1}
     \cmidrule(r){2-3}
     \cmidrule(){4-6}
    Predicted & 69.0 & 73.8 & 71.5 & 30.0 & 57.7\\
    Max-size        & 92.8 & 80.0 & 87.9 & 73.0 & 62.8 \\
    \bottomrule
     \end{tabular}
    \end{center}
    \vspace{\captionSpace}
    \caption{\textbf{Assigned proposal cover rate and consistency.} 
    Left: ratio of assigned proposal covering the ground truth both. We evaluate on an ImageNet subset that has box ground truth and the annotated COCO training set; Right: average assigned bounding box IoU of between the final model and the half-schedule model.}
    \lbltab{analysis}
    \end{subtable}
\end{center}
\vspace{-5mm}
\caption{\textbf{Comparison between predicted loss and and max-size loss.} 
\textbf{(a)}: comparison under different baselines.
\textbf{(b)}: comparison in customized metrics.}
\vspace{-10mm}
\end{table}

Our non-prediction-based losses perform significantly better than prediction-based losses as is shown in ~\reftab{imagelabel}.
In this section, we take the max-size loss and the predicted-loss as the representitives and conduct more detailed comparisons between them.
A straightforward reason is that the predicted loss
requires a good initial prediction to guide the pseudo-label-based training.
However in the \zeroshot{} detection setting the initial predictions are inherently flawed.
To verify this, in ~\reftbl{max-score-gap}, we show both improving the backbone and including rare classes in training can narrow the gap.
However in the current performance regime, our max-size loss performs better.

We highlight two additional advantages of the max-size loss that may contribute to the good performance:
(1) the max-size loss is a safe approximation of object regions;
(2) the max-size loss is consistent during training.
\reffig{qualitative-proposal} provides qualitative examples of the assigned region for the predicted loss and the max-size loss.
First, we observe that while being coarse at the boundary, the max-size loss can \emph{cover} the target object in most cases.
Second, the assigned regions of the predicted loss are usually different across training iterations, especially in the early phase where the model predictions are unstable.
On the contrary, max-size loss supervises consistent regions across training iterations.

\reftab{analysis} quantitatively evaluates these two properties.
We use the ground truth box annotation in the full COCO detection dataset and a subset of ImageNet with bounding box annotation \footnote{\url{https://image-net.org/download-bboxes.php}. 213K of the 1.2M IN-L images have bounding box annotations.} to evaluate the cover rate.
We define cover rate as the ratio of image labels whose ground-truth box has $>0.5$ intersection-over-area with the assigned region.
We define the consistency metric as the average assigned-region IoU of the same image between the $1/2$ schedule and the final schedule.
\reftab{analysis} shows max-size loss is more favorable than predicted loss on these two metrics.
However we highlight that these two metrics alone do not always correlate to the final performance, as the \textbf{image-box} loss is perfect on both metrics but underperforms max-size loss.

\section{ViLD baseline details}
\lblsec{vild-details}

The baseline in ViLD~\cite{gu2021zero} is very different from detectron2.
They use MaskRCNN detector~\cite{He_2017_ICCV} with Res50-FPN backbone, but trains the network
from scratch without ImageNet pretraining.
They use large-scale jittering~\cite{ghiasi2021simple} with input resolution $1024\times1024$ and train a $32\times$ schedule. 
The optimizer is SGD with batch size $256$ and learning rate $0.32$.
We first reproduce their baselines (both the oracle detector and ViLD-text) under the same setting.
We observe half of their schedule ($16\times$) is sufficient to closely match their numbers.
The half training schedule takes $4$ days on $4$ nodes (each with 8 V100 GPUs).
We then finetune another $16\times$ schedule using ImageNet data with our max-size loss.

\section{Open-vocabulary COCO benchmark details}
\lblsec{coco-details}

Open-vocabulary COCO is proposed by Bansal et al.~\cite{bansal2018zero}.
They manually select 48 classes from the 80 COCO classes as base classes,
and 17 classes as novel classes.
The training set is the same as the full COCO, but only images containing at least one base class are used.
During testing, we report results under the ``generalized zero-shot detection''
setting~\cite{bansal2018zero}, where all COCO validation images are used.

We strictly follow the literatures~\cite{bansal2018zero,rahman2020improved,zareian2021open}
 to use FasterRCNN~\cite{ren2015faster} with ResNet50-C4 backbone and the $1\times$ training schedule ($90k$ iterations).
We use horizontal flip as the only data augmentation in training and 
keep the input resolution
fixed to $800\times1333$ in both training and testing.
We use SGD optimizer with a learning rate $0.02$ (dropped by $10\times$ at $60k$ and $80k$
iteration) and batch size $16$.
The evaluation metric on open-vocabulary COCO is box mAP at IoU threshold 0.5.
Our reproduced baseline matches OVR-CNN~\cite{zareian2021open}.
Our model is finetuned on the baseline model with another $1\times$ schedule. We sample detection data and image-supervised data in a $1:1$ ratio.

\reftab{imagelabel-coco} repeats the experiments in \reftab{imagelabel} on \zeroshot{} COCO. The observations are consistent: our proposed non-prediction-based methods outperform existing prediction-based counterparts, and the max-size loss performs the best among our variants.

\begin{table*}[!t]
    \begin{center}
    \begin{tabular}{@{}ll@{\ \ \ \ }c@{\ \ \ \ \ \ }c@{}}
    \toprule
    & & mAP50$^{\text{box}}_{\text{all}}$ & mAP50$^{\text{box}}_{\text{novel}}$ \\
    \midrule
    \rowNumber{1} & Box-Supervised (base cls) & 39.3 & 1.3 \\
    \midrule
    \rowNumber{2} & Self-training~\cite{sohn2020simple} & 39.5 & 1.8\\
    \rowNumber{3} & WSDDN~\cite{bilen2016weakly} & 39.9 & 5.9 \\
    \rowNumber{4} & DLWL*~\cite{ramanathan2020dlwl} & 42.9 & 19.6 \\
    \rowNumber{5} &
    Predicted~\cite{redmon2017yolo9000}
        & 41.9 & 18.7\\
    \midrule
    \rowNumber{6} & \OURS{} (Max-object-score) & 43.3 & 20.4 \\
    \rowNumber{7} & \OURS{} (Image-box) & 43.4 & 21.0\\
    \rowNumber{8} & \OURS{} (Max-size) & \bf 44.7 & \bf 24.1\\
    \midrule
    \rowNumber{9} & \color{gray} Box-Supervised (all cls)
      & \color{gray} 54.9 & \color{gray} 60.0 \\
    \bottomrule
     \end{tabular}
    \end{center}
    \vspace{\captionSpace}
    \caption{
    \textbf{Different ways to use image supervision on \zeroshot{} COCO.}
    The models are trained using the OVR-CNN~\cite{zareian2021open} recipe with ResNet50-C4~\cite{bansal2018zero} backbone.
    We follow setups in~\reftab{imagelabel}. The observations are consistent with LVIS.
    }
    \lbltab{imagelabel-coco}
    \vspace{-5mm}
\end{table*}

\section{Compare to MosaicOS~\cite{zhang2021mosaicos}}

\lblsec{full-lvis}

\lblsec{mosaic-details}
\begin{table}[!t]
\begin{center}
\begin{tabular}{@{}l@{\ \ \ }c@{\ \ \ }c@{\ \ \ }}
\toprule 
 & mAP$^\text{mask}$ & mAP$^\text{mask}_{\text{r}}$ \\
\midrule
Box-Supervised~\cite{zhang2021mosaicos} & 22.6 & 12.3 \\
MosaicOS~\cite{zhang2021mosaicos} & 24.5 \scriptsize \textcolor{green4}{(+1.9)} & 18.3  \scriptsize \textcolor{green4}{(+6.0)} \\
\midrule
Box-Supervised (Reproduced) & 22.6 & 12.3 \\
\OURS{} (default classifier) &  25.1 \scriptsize \textcolor{green4}{(+2.5)} & 18.6 \scriptsize \textcolor{green4}{(+6.3)} \\
\midrule
Box-Supervised (CLIP classifier) & 22.3 & 14.1 \\
\OURS{} (CLIP classifier) & \bf 24.9 \scriptsize \textcolor{green4}{(+2.6)} & \bf 20.7 \scriptsize \textcolor{green4}{(+6.5)} \\
\bottomrule
 \end{tabular}
\end{center}
\vspace{\captionSpace}
\caption{
    \textbf{Standard \lvis{} compared to MosiacOS~\cite{zhang2021mosaicos}.}
    Top block: results quoted from MosiacOS paper; Middle block: \OURS{} with the default random intialized and trained classifier; Bottom block: \OURS{} with CLIP classifier.}
\lbltab{mosiacos}
\vspace{-5mm}
\end{table}

MosaicOS~\cite{zhang2021mosaicos} first uses image-level annotations to improve LVIS detectors.
We compare to MosaicOS~\cite{zhang2021mosaicos} by strictly following their baseline setup (without any improvements in~\cref{sec:implementation_details}).
The detailed hyper-parameters follow the detectron2 baseline as described in ~\cref{sec:lvis-baseline}.
We finetune on the Box-supervised model with an additional $2\times$ schedule with Adam optimizer.
~\reftab{mosiacos} shows our re-trained baseline exactly matches their reported results from the paper.
Our method is developed based on the CLIP classifier, and we also report our baseline with CLIP.
The baseline has slightly lower mAP and higher mAP$_{\text{r}}$.
MosaicOS uses \imnetLvis{} and additional web-search images as image-supervised data.
\OURS{} outperforms MosaicOS~\cite{zhang2021mosaicos} in mAP and \mAPr,
without using their multi-stage training and mosaic augmentation.
Our relative improvements over the baseline are slightly higher than MosiacOS~\cite{zhang2021mosaicos}.
We highlight our training framework is simpler and we use less additional training data (Google-searched images).

\section{Generalization to Deformable-DETR.}
\lblsec{detr}
\begin{table}[!t]
    \begin{center}
    \begin{tabular}{@{}l@{\ }c@{\ }c@{\ }c@{\ }c@{\ }@{}}
    \toprule
      & mAP$^{\text{box}}$ & mAP$^{\text{box}}_{\text{r}}$ & mAP$^{\text{box}}_{\text{c}}$ & mAP$^{\text{box}}_{\text{f}}$ \\
    \midrule
    \baselineDet & 31.7 & 21.4 & 30.7 & \bf 37.5 \\
    \OURS{} & \bf 32.5 & \bf 26.2 & \bf 31.3 & 36.6 \\
    \bottomrule
     \end{tabular}
    \end{center}
    \vspace{\captionSpace}
    \caption{
    \textbf{\OURS{} applied to Deformable-DETR~\cite{zhu2020deformable}.}
    We report Box mAP on full LVIS. Our method improves Deformable-DETR.}
    \vspace{-8mm}
    \lbltab{detr}
    \end{table}
    
We apply \OURS{} to the recent Transformer based Deformable-DETR~\cite{zhu2020deformable} to study its generalization.
We use their default training recipe, Federated Loss~\cite{zhou2021probabilistic} and train for a $4\times$ schedule ($\sim\!48$ LVIS epochs).
We apply the image supervision to the query from the encoder with the max predicted size.
~\reftab{detr} shows that \OURS{} improves over the baseline (+$0.8$ mAP and $+4.8$ mAP$_{\text{r}}$) and generalizes to Transformer based detectors.

\begin{table}[!b]
    \vspace{-5mm}
    \begin{center}
    \begin{tabular}{@{}l@{\ \ \ \ \ \ }c@{\ \ \ \ \ \ }c@{\ \ \ \ \ \ }c@{}}
    \toprule 
      & mAP$^\text{mask}$ & mAP$^\text{mask}_{\text{IN-L}}$ & mAP$^\text{mask}_{\text{non-IN-L}}$ \\
    \midrule
    Box-Supervised & 30.2 & 30.6 & 27.6 \\
    Max-size & 32.4 & 33.5 & 28.1\\
    \midrule
    % \midrule
    & mAP$^\text{mask}$ & mAP$^\text{mask}_{\text{CC}}$ & mAP$^\text{mask}_{\text{non-CC}}$ \\
    \midrule
    Box-Supervised & 30.2 & 30.1 & 29.5 \\
    Max-size & 30.9 & 31.7 & 28.6 \\
    \bottomrule
     \end{tabular}
    \end{center}
    \vspace{\captionSpace}
    \caption{\textbf{mAP breakdown into classes with and without image labels.}
    Top: \OURS{} trained on ImageNet. Bottom: \OURS{} trained on CC. Most of the improvements are from classes with image-level labels. On ImageNet \OURS{} also improves classes without image labels thanks to the CLIP classifier.}
    \lbltab{breakdown}
    % \vspace{-3mm}
\end{table}

\section{Improvements breakdown to classes}
\lblsec{mAp-fixed}

\reftab{breakdown} shows mAP breakdown into classes with and without image labels for both the Box-Supervised baseline and \OURS{}. As expected, most of the improvements are from classes with image-level labels. On ImageNet, \OURS{} also improves classes without image labels thanks to the CLIP classifier which leverages inter-class relations.

\section{mAP$^{\text{Fixed}}$ evaluation}
\lblsec{mAp-fixed}

\begin{table}[!t]
% \vspace{-5mm}
\begin{center}
\begin{tabular}{@{}l@{\ }c@{\ }c@{\ }c@{\ }c@{}c@{}}
\toprule 
Datasets & mAP$^\text{box}$ & mAP$^\text{box}_{\text{novel}}$ & mAP$^{\text{Fixed}}$ & mAP$_{\text{novel}}^{\text{Fixed}}$ \\
\cmidrule(r){1-1}
\cmidrule(r){2-3}
\cmidrule(r){4-5}
\baselineDet & 30.2 & 16.4 & 31.2 & 18.2 \\
\OURS{} & 32.4  \scriptsize \textcolor{green4}{(+2.2)} & 24.9  \scriptsize \textcolor{green4}{(+8.5)} & 33.4  \scriptsize \textcolor{green4}{(+2.3)} & 26.7  \scriptsize \textcolor{green4}{(+8.5)} \\
\bottomrule
    \end{tabular}
\end{center}
\vspace{\captionSpace}
\caption{\textbf{mAP$^{\text{Fixed}}$ evaluation}. Middle: the original box mAP metric used in the main paper. Right: the new box mAP$^{\text{Fix}}$ metric. Our improvements are consistent under the new metric.}
\lbltab{mAPfixed}
\vspace{-4mm}
\end{table}

\reftab{mAPfixed} compares our improvements under the new mAP$^{\text{fix}}$ proposed 
in Dave \etal~\cite{dave2021evaluating}. Our improvements are consistent under the new metric.

\section{Image Attributions}
License for the images from OpenImages in ~\reffig{qualitative}:
\small
\begin{itemize}
\item ``Oyster'': Photo by The Local People Photo Archive (CC BY 2.0)
\item ``Cheetah'': Photo by Michael Gil (CC BY 2.0)
\item ``Harbor seal'': Photo by Alden Chadwick (CC BY 2.0)
\item ``Dinosaur'': Photo by Paxson Woelber (CC BY 2.0)
\end{itemize}

\end{document}